\documentclass[preprint]{elsarticle}

\advance\textwidth by 80pt

\advance\oddsidemargin by -40pt

\advance\evensidemargin by -40pt

\advance\textheight by 40pt

\advance\topmargin by -40pt

\usepackage[dvips]{epsfig}
\usepackage{epic}
\usepackage{array}
\usepackage{epstopdf}
\usepackage{booktabs}
\usepackage{a4wide}
\usepackage{verbatim}
\usepackage{amsmath}
\usepackage{amssymb}
\usepackage{amsfonts}
\usepackage{amstext}
\usepackage{amsbsy}
\usepackage{graphicx}
\usepackage[symbol]{footmisc}
\usepackage{subfigure}
\usepackage{longtable}
\usepackage{fancybox}
\usepackage{fancyvrb}
\usepackage{amsthm}
\usepackage{xcolor}
\usepackage{url}
\usepackage{lineno}
\usepackage{natbib}

\newcommand{\beq}{\begin{equation}}
\newcommand{\eeq}{\end{equation}}

\newcommand{\bX}{\mathbf{X}}
\newcommand{\bW}{\mathbf{W}}
\newcommand{\bM}{\mathbf{M}}

\newcommand{\bT}{\mathbf{T}}
\newcommand{\bE}{\mathbf{E}}
\newcommand{\bP}{\mathbf{P}}

\newcommand{\bA}{\mathbf{A}}

\newcommand{\bI}{\mathbf{I}}

\newcommand{\trc}{{\rm tr}}
\newcommand{\tT}{{\rm T}}

\newcommand{\diag}{{\rm diag}}

\newlength{\uvafigsize}
\DeclareMathOperator*{\argmax}{arg\,max}

\DeclareMathOperator*{\argmin}{arg\,min}

\begin{document}
	\begin{frontmatter}

\title{All Sparse PCA Models Are Wrong, But Some Are Useful. \\Part I: Computation of Scores, Residuals and Explained Variance} 

\author{J. Camacho \fnref{fn1}}
\author{A. K. Smilde \fnref{fn2}}
\author{E. Saccenti \fnref{fn3}}
\author{J. A. Westerhuis \fnref{fn2}}
\fntext[fn1]{Department of Signal Theory, Telematics and Communications, School of Computer Science and Telecommunications - CITIC,	University of Granada, Granada, Spain}
\fntext[fn2]{Biosystems Data Analysis, University of Amsterdam, Amsterdam, the Netherlands}
\fntext[fn3]{Laboratory of Systems and Synthetic Biology, Wageningen University \& Research, Wageningen, the Netherlands}
\cortext[cor1]{Corresponding author: J. Camacho josecamacho@ugr.es} 

\date{}

\begin{abstract}
Sparse Principal Component Analysis (sPCA) is a {popular} matrix factorization approach based on Principal Component Analysis (PCA) that combines variance maximization and sparsity with the ultimate goal of improving data interpretation. {When moving from PCA to sPCA, there are a number of implications that the practitioner needs to be aware of. A relevant one is that scores and loadings in sPCA may not be orthogonal. For this reason, the traditional way of computing scores, residuals and variance explained that is used in the classical PCA cannot directly be applied to sPCA models. This also affects how sPCA components should be visualized. 
In this paper we illustrate this problem both theoretically and numerically using simulations for several state-of-the-art sPCA algorithms, and provide proper computation of the different elements mentioned. {We show that sPCA approaches present disparate and limited performance when modeling noise-free, sparse data. In a follow-up paper, we discuss the theoretical properties that lead to this problem. }

} 
\end{abstract}

\end{frontmatter}

\section{Introduction}

In modern data analysis there is an increased interest in sparse (component) methods to obtain models that are simpler and easier to interpret. One of such methods is the popular sparse Principal Component Analysis (sPCA) \cite{Jolliffe2003,Zou2006,PMD,GPCA}. While the properties of sPCA have already been discussed  \cite{PMD,Mackey2008,Hastie:2015:SLS:2834535},  there are several aspects that have not been treated in detail and that are relevant for the interpretation of the models. 

{sPCA is based on Principal Component Analysis (PCA) \cite{Jolliffe02, Jackson03}, which factorizes a data matrix using the criterion of maximizing variance. The PCA model follows the expression:
\begin{equation} \label{eq:PCAm}
\mathbf{X} = \mathbf{\hat{T}}\mathbf{\hat{P}}^\tT +
\mathbf{E},
\end{equation}
where $\mathbf{X}$ is a $N \times M$ data matrix,
$\mathbf{\hat{T}}$ the $N \times A$ score matrix, $\mathbf{\hat{P}}$ the $M \times A$ loading matrix and $\mathbf{E}$ the $N \times M$ matrix of residuals. {An interesting property of PCA is that the loadings and the scores are orthogonal and, as a result, they can be computed either using a simultaneous or sequential algorithm.}

When moving from PCA to sPCA, there are at least two interrelated  consequences that need to be considered: {\textit{i}) the appearance of correlation among scores and/or loadings in different components and \textit{ii}) the departure of components from the row-space defined by the original data. The first point makes the traditional way of computing scores, residuals and captured variance used in PCA not applicable in sPCA. Moreover, the captured variance is often used as a criterion for model quality and for selection/comparison of model variants. Therefore, its accurate computation is necessary. The second point affects interpretability, specially for multi-component models. 
Both effects are especially relevant in those domains in which the interpretation of the scores/residuals is necessary, like in chemometrics.} 

Zou  \textit{et al.} \cite{Zou2006} proposed a method based on the QR decomposition to compute the variance captured by sPCA when scores are correlated; we will show that this method is generally incorrect. Witten \textit{et  al.} \cite{PMD} related the correlation of the loadings/scores to the departure from the row/column-space of the data and proposed an orthogonalized version of their sPCA method, where scores are made orthogonal; we will show that this orthogonalization reduces the captured variance and dwarfs interpretation, {but improves the estimation of sparse loadings}. Mackey \cite{Mackey2008} discussed different deflation methods in the context of sPCA, some of which compute residuals incorrectly, leading to the double-counting of variance, \textit{i.e.}, a portion of variance in the data captured by more than one component. He also contributed a new deflation method which, we will show, still requires the corrections that will be proposed and discussed in this paper. 

In this study we present a comparison among several widely used sPCA algorithms and reflect on how scores, residuals and captured variance should be computed for sparse components, something central in chemometrics. The problem of the departure of components from the row-space is treated in a follow-up paper. 
The rest of the paper is organized as follows. In Section 2, we review a set of sPCA algorithms considered. In Section 3, we discuss the mathematical properties of sPCA, using PCA as a baseline. In Section 4, we propose two simulated examples that mimic spectral data, and a Monte Carlo experiment, that will illustrate interpretation problems and solutions in the following sections. Section 5 discusses the implications of correlated scores and loadings in sPCA. Section 6 presents the correct computation of scores, residuals and captured variance in sPCA models. Section 7 offers some final considerations and conclusions.

\section{Sparse PCA algorithms}

{Most sPCA algorithms modify the classical PCA by including sparse-inducing constraints or penalties with the $L_0$ or $L_1$ norms \cite{Richt2012}. Jolliffe  \textit{et al.} \cite{Jolliffe2003} developed the SCoTLASS algorithm (Simplified Component Technique-LASSO), which incorporates the least absolute shrinkage and selection operator (lasso) \citep{Tibshirani1994}, where the $L_1$ norm (absolute values) of the loadings is penalized.} The SCoTLASS criterion follows:

\begin{equation} \label{lasso}
\mathbf{\hat{p}}^{SL} = \argmax_{\mathbf{p}} \| \mathbf{X} \mathbf{p} \|_F^2 \quad s.t. \,\,\,  \|\mathbf{p}\|_1 \le c, \quad \|\mathbf{p}\|_2^2 = 1
\end{equation}

\noindent where $\mathbf{\hat{p}}^{SL}$ is the resulting sparse loading, with superscript $SL$ referring to SCoTLASS. To obtain successive components, the SCoTLASS optimization constraints the 2nd and further sparse loadings to be orthogonal to the rest. The SCoTLASS criterion is very computational demanding \cite{Hastie:2015:SLS:2834535}, and this makes its use unpractical for data exploration. In addition, the number of nonzero elements in $\mathbf{p}^{SL}$ is upper-bounded by the number of observations in the data, which is a structural limitation of the lasso constraint. 

\subsection{Sparse PCA by Zou, Hastie and Tibshirani}

Zou \textit{et al.} \cite{Zou2006} introduced an alternative formulation to generate sparse components, motivated by the structural limitation of the {lasso} and the computation cost of SCoTLASS. This new model was also named Sparse PCA (SPCA, we will use a capital 'S' to differentiate this approach from the general sPCA technique, {although at that time the term was already used to designate the group of sparse variants of PCA}. SPCA redefines PCA as a regularized regression problem with the ridge penalty, where the responses are the PCA scores. In this formulation any loading vector of PCA, $\mathbf{\hat{p}}$, equals $\frac{\mathbf{\hat{p}}^{R}}{\|\mathbf{\hat{p}}^{R}\|}$, where
\begin{equation} \label{PCA}
\mathbf{\hat{p}}^{R} = \argmin_{\mathbf{p}} \| \mathbf{\hat{t}} - \mathbf{X} \mathbf{p} \|_F^2 + \lambda_2 \|\mathbf{p}\|^2_2
\end{equation}
\noindent with superscript $R$ referring to \emph{regression}, and $\mathbf{\hat{t}}$ is the corresponding score vector. {Therefore, $\mathbf{\hat{p}}^{R} \propto \mathbf{\hat{p}}$, and $\mathbf{\hat{p}}$ is obtained by normalizing $\mathbf{\hat{p}}^{R}$ to unit length}. This equivalence holds in general for any non-negative $\lambda_2$. {{Equivalently, the loadings of PCA can be obtained from the following regression problem followed by a normalization step:}
\begin{equation} \label{PCA2}
\{\mathbf{\hat{P}}^R, \mathbf{\hat{Q}}^{R}\} = \argmin_{\mathbf{P},\mathbf{Q}} \| \textbf{X} - \mathbf{X} \mathbf{P} \mathbf{Q}^\tT\|_F^2 + \lambda_2  \sum_{a=1}^A \|\mathbf{p}_a\|^2_2 \quad s.t.\quad  \mathbf{Q}^\tT\mathbf{Q} = \mathbf{I}
\end{equation}

\noindent {where $\mathbf{p}_a$ represents the $a$th column vector in $\mathbf{P}$, with $A$ the number of components, and we distinguish between weights $\mathbf{\hat{P}}^R$ and loadings $\mathbf{\hat{Q}}^R$. In similar modeling methods, like Partial Least Squares, weights are used for the derivation of scores from the data and loadings for the identification of the captured variance in $\textbf{X}$, e.g. for deflation. The advantage of Eq. (\ref{PCA2}) over Eq. (\ref{PCA}) is that the former does not require the pre-computation of the PCA scores. 	
$\mathbf{\hat{P}}^R$ is proportional to the PCA loadings for any non-negative $\lambda_2$. Therefore, we can understand Eq. (\ref{PCA2}) as a flexible method to compute PCA, where constraints and penalties can be introduced either in the weights or the loadings. }

Using this flexible redefinition of PCA the {lasso} can be implemented using a criterion close to the (naive) elastic net \citep{Zou2005}, which is a combination of both the {lasso} and the ridge constraint:
\begin{equation} \label{SPCA}
\{\mathbf{\hat{P}}^{SP}, \mathbf{\hat{Q}}^{SP}\} = \argmin_{\mathbf{P},\mathbf{Q}} \| \textbf{X} - \mathbf{X} \mathbf{P} \mathbf{Q}^\tT\|_F^2 + \lambda_2 \sum_{a=1}^A \|\mathbf{p}_a\|^2_2 + \lambda_1 \sum_{a=1}^A  \|\mathbf{p}_a\|_1 \quad s.t.\quad  \mathbf{Q}^\tT\mathbf{Q} = \mathbf{I}
\end{equation}

\noindent with superscript $SP$ referring to \emph{SPCA}. {In the following, we will refer to $\mathbf{\hat{P}}^{SP}$ as sparse loadings and to $\mathbf{\hat{Q}}^{SP}$ as auxiliary loadings to simplify the discussion in connection with other sPCA techniques. Note that the roles of $\mathbf{\hat{P}}^R$ and $\mathbf{\hat{Q}}^R$ are performed by a unique set of loadings in PCA and most sPCA variants.}
The SPCA algorithm is simultaneous, in the sense that all components are identified in the same {run}. 
%
{Note the expression in Eq. (\ref{SPCA}) is not optimal to approximate data, {because $\mathbf{{Q}}$ is constrained to be orthonormal, while at the same time the norms of the vectors in $\mathbf{{P}}$ are penalized}, leaving no flexibility to approximate $\mathbf{X}$ and resulting in very large residuals. Besides, the sparse loadings are not normalized. The authors suggest to normalize them after the optimization, in the last step of their algorithm, but they overlook the need to update $\mathbf{\hat{Q}}$ accordingly. This reflects the fact that authors were only interested in  $\mathbf{\hat{P}}^{SP}$ as the output of the algorithm, and the model in Eq. (\ref{SPCA}) is only a means to arrive to such output.}

They also realized that this SPCA algorithm produces highly correlated scores, and concluded that the current method for computing the captured variance as $\trc(\mathbf{\hat{T}}^\tT\mathbf{\hat{T}})$ used in PCA is too optimistic. For this reason, they suggested to use the QR decomposition of $\mathbf{\hat{T}}$, which is $\mathbf{\hat{T}} = \mathbf{Q}_T\mathbf{R}_T$ for $\mathbf{Q}_T$ orthogonal and $\mathbf{R}_T$ upper triangular, and to compute the captured variance of the model as the sum of squares of the elements in the diagonal of $\mathbf{R}_T$. 

An alternative SPCA algorithm has been published as part of the SPASM Toolbox \cite{Sjostrand2012}, which is a sequential variant of SPCA, in which the loadings are computed one at a time. 
This algorithm is faster than the simultaneous versions, with a small reduction of captured variance. 

%
%


\subsection{The Penalized Matrix Decomposition  by Witten, Tibshirani and Hastie} \label{PMDsec}

Witten \textit{et al.} \cite{PMD} introduced a new iterative algorithm based on alternating of cardinality-constrained rank-one variance maximization and matrix deflation. For the first task, they proposed the Penalized Matrix Decomposition (PMD), a biconvex problem solved by alternating between the identification of {$\mathbf{u}$ (left pseudo-eigenvector) and $\mathbf{p}$ (right pseudo-eigenvector)}. PMD equals the 1-rank Singular Value Decomposition (SVD) when no constrains are considered. The sPCA formulation in the PMD framework is:

\begin{equation} \label{PMD}
\{\mathbf{\hat{p}}^{P}, \mathbf{\hat{u}}^{P}\} = \argmax_{\mathbf{p},\mathbf{u}}\  \mathbf{u}^\tT \mathbf{X} \mathbf{p}   \quad s.t. \quad \|\mathbf{p}\|_1 \le c_2, \quad \|\mathbf{p}\|_2^2 \le 1, \quad \|\mathbf{u}\|_2^2 \le 1
\end{equation}
\noindent {where $\mathbf{\hat{p}}^{P}$} is obtained using a soft-thresholding operator, {with superscript $P$ referring to \emph{PMD}}. The corresponding pseudo-singular value is obtained as:

\begin{equation} \label{PMD2}
\hat{d}^{P} = (\mathbf{\hat{u}}^{P})^\tT \mathbf{X} \mathbf{\hat{p}}^{P}
\end{equation}

{After each component is obtained, deflation is performed as:
	
	\begin{equation} \label{pd}
	\mathbf{X} = \mathbf{X}  - \hat{d}^{P} \mathbf{\hat{u}}^{P}(\mathbf{\hat{p}}^{P})^\tT
	\end{equation}}

This solution is related to SCoTLASS and sPCA \cite{PMD}, but according to \cite{Hastie:2015:SLS:2834535} this method is more efficient, something we have also seen experimentally. Regarding the deflation, they state that while projection deflation works well for true eigenvectors, when using constrained versions of PCA (\textit{i.e.} sPCA) $\mathbf{u}$ and/or $\mathbf{p}$ are not in the column and row spaces of $\mathbf{X}$. 
As an alternative to projection deflation, they propose to compute  $\mathbf{u}$ using an orthogonalized factor to all the previous left pseudo-eigenvectors. By incorporating this deflation, orthogonal scores are obtained. However, the authors themselves claim that it is not clear whether orthogonality is a desirable property \cite{PMD,Hastie:2015:SLS:2834535}. We will call this the orthogonalized deflation approach. 


\subsection{Group-wise Principal Component Analysis} \label{GPCAsec}

The Group-wise Principal Component Analysis (GPCA) \cite{GPCA} is a sparse approach to PCA proposed by some of the authors. The main difference between GPCA and the sPCA variants presented in the previous sections is that sparsity is not achieved by using a loss function with the LASSO constraint but it is defined in terms of groups of correlated variables and it is imposed using a set of nested PCA loops. Every component contains non-zero loadings for a single group of correlated variables, and this component is calibrated using only this subset of variables.\\
\indent GPCA starts with the identification of a set of $G$ (possibly overlapping) groups of correlated variables obtained from a map $\bM$, with elements $m_{ij} \in [-1, 1]$ containing the strength of the relationship between variables $i$ and $j$. An example of this map is the correlation matrix of $\bX$. In the original formulation of GPCA, the MEDA approach (Mis\-sing-data for Exploratory Data analysis) \cite{Camacho2011missing} was implemented to define $\bM$, which  uses a missing data strategy to estimate the correlation between any two variables, an approach which has been found to be effective in filtering out noise when estimating correlations \cite{Camacho2011missing}.

Once the $G$ groups have been defined, the GPCA algorithm computes $G$ candidate loading vectors, where the $g$-th vector has non zero elements associated to the variables in the $g$-th group. From these, only the loading with the largest variance is retained and it is used to deflate data matrix $\bX$.  The algorithm iterates until a set of $A$ pseudo-components is extracted. 

{In GPCA, the deflation by Mackey \cite{Mackey2008} is used} who showed that the common approach of single-component deflation in PCA is in general not valid for sPCA. This is the projection deflation, equivalent to Eq. (\ref{pd}):
	
	\begin{equation}\label{XAres}
	\mathbf{X}_A = \mathbf{X}_{A-1} (\mathbf{I} -  \mathbf{\hat{p}_A}(\mathbf{\hat{p}_A}^\tT \mathbf{\hat{p}_A})^{-1}\mathbf{\hat{p}_A}^\tT)
	\end{equation}
where $\mathbf{X}_{A-1}$ denote the residuals after $A-1$ PCs have been extracted. After normalizing $\mathbf{\hat{p}_A}$ to unit length, Eq. (\ref{XAres}) simplifies to:
	
	\begin{equation}
	\mathbf{X}_A = \mathbf{X}_{A-1}  (\mathbf{I} -  \mathbf{\hat{p}_A}\mathbf{\hat{p}_A}^\tT)
	\end{equation}

Setting $\mathbf{XX}_{A-1}$ to be the covariance matrix of a data sample $\mathbf{X}$ after $A-1$ PCs have been extracted, the projection deflation can be written as
	
	\begin{equation}
	\mathbf{XX}_{A} = (\mathbf{I} - \mathbf{\hat{p}}_A \mathbf{\hat{p}}^\tT_A) \mathbf{XX}_{A-1} (\mathbf{I} - \mathbf{\hat{p}}_A \mathbf{\hat{p}}^\tT_A)
	\end{equation}

For $\mathbf{\hat{p}_A}$ a pseudo-eigenvector of $\mathbf{XX}_{A-1}$, 
Mackey \cite{Mackey2008} shows that projection deflation annihilates the corresponding pseudo-eigenvalue while maintaining the others in the data. 
However, when deflating with respect to a series of non-orthogonal pseudo-eigenvectors, projection deflation may lead to re-introduce previously deflated components and to the "double counting" of variance, which can cause misleading results for interpretation. Mackey also shows that this problem can be solved by annihilating the variance captured by the pseudo-eigenvector that is orthogonal to the space spanned by the previous extracted pseudo-eigenvectors. For that, he introduces a combination of projection deflation and orthogonalized deflation, named orthogonalized projection deflation. This, similarly to \cite{Zou2006}, adds to the sparse loadings a set of auxiliary orthogonal right pseudo-eigenvectors to perform the deflation.

One way to do this deflation follows a sequential Gram-Schmidt decomposition:

\begin{equation}
\mathbf{\hat{q}}_{A} = \mathbf{B}_{A-1} \mathbf{\hat{p}}_{A}
\end{equation}
\begin{equation}
\mathbf{X}_{A} = \mathbf{X}_{A-1} (\mathbf{I} - \mathbf{\hat{q}}_{A} \mathbf{\hat{q}}_{A}^\tT)
\end{equation}
\begin{equation}
\mathbf{B}_{A} = \mathbf{B}_{A-1} (\mathbf{I} - \mathbf{\hat{q}}_{A} \mathbf{\hat{q}}_{A}^\tT)
\end{equation}

\noindent where  $\mathbf{B}_{0}$ is initialized to the identity matrix. We will call this approach the Mackey's generalized deflation.


\section{From principal components to sparse principal components}

When moving from a classical principal component model to any of it sparse variants there are a number of implications that we think is pivotal to {assess, especially in the context of  data interpretation}:

\begin{enumerate}
	
	\item \textbf{Correlation of scores}\\
	{The scores resulting from a standard PCA are uncorrelated as a consequence of maximizing variance {in each component}, while the scores of sPCA model are typically correlated. This correlation complicates proper visualization but also allows for a more flexible modeling. As already discussed, Witten \textit{et al.} \cite{PMD} provide a solution to correct for this correlation, but it is not clear whether it should be used.} More importantly, care should be taken when computing the captured variance with correlated scores.
	\item \textbf{Correlation of loadings}\\
	Correlation of loadings is a relevant problem, in particular when scatter plots of scores are used for interpretation. The visualization in scatter plots assumes orthogonal axes in the original variable space. This holds for the standard PCA components, but it does not necessarily hold for sparse components. {Again, care should be taken when computing the captured variance with correlated loadings.}
	
	\item \textbf{Loss of captured variance per principal component}.\\
	The PCs are {optimal} in terms of captured variance. This is a property that makes them especially useful for data interpretation, since a large portion of variance/information is explained by a reduced set of {components. This in turn simplifies the visualization of the data, since a minimum number of PCs needs to be visualized to account for a certain level of variance}. However, part of the captured variance might not be of interest and could even complicate interpretation. As a relevant example, all PCs include a certain amount of embedded noise \cite{malinowski2002factor}. When we move to the sparse context, we simplify the interpretation of loadings, but also need more components to capture the same amount of variance. 
	
	\item \textbf{Moving outside the row-space}\\
	{Sparse loadings can be outside the data row-space as a result of constraints/penalties applied. As a matter of fact, there is no sparse solution inside the row-space in the presence of noise. 
	If the constraints/penalties are reasonable, we can improve the model quality with this departure. A notable example of this are the non-negative constraints, useful to model non-negative data. While the departure of a sparse component can be advantageous, it also complicates the generation of multi-component models. 
}
		
	
\end{enumerate}
	
We will discuss these implications in detail and quantify them using simulation in the next Sections, except for the departure from the row-space, which will be treated in the follow-up paper.}
	
\section{Simulated examples and Monte Carlo study}
	
{We will use a number of examples with simulated data to support our discussion,  
	{where we will opt not to introduce noise so that we can easily illustrate the behavior of sPCA algorithms.} First, we will simulate two different examples for visual illustration, representing two common situations in exploratory data analysis that are of interest for this study: when orthogonality is consistent with the underlying data generation process and when it is not. The motivation is that a subset of sPCA methods and PCA impose orthogonality in some or all the factors, and we want to assess to which extent this is a useful constraint.  As a second step we will generalize the results through a Monte Carlo simulation. 
We will consider several variants and implementations of sPCA, some of which are evaluated and compared in this paper for the first time:
		
		\begin{enumerate}
			\item The sPCA algorithm by Zou \textit{et al.} \cite{Zou2006} (SPCA) and the sequential (iterative) implementation in the SPASM toolbox \cite{Sjostrand2012} (SPCA-Sq).
			\item The PMD algorithm by \cite{PMD} with projection (PMD-PD) and orthogonalized deflation (PMD-O) discussed in section \ref{PMDsec}, and a modified version using Mackey's \cite{Mackey2008} generalized deflation (PMD-M) discussed in section \ref{GPCAsec}.
			\item The GPCA algorithm by \cite{GPCA} using projection deflation (GPCA-PD), Mackey 's \cite{Mackey2008} generalized deflation (GPCA-M) and orthogonalized deflation by \cite{PMD} (GPCA-O). 
		\end{enumerate}

\subsection{Orthogonal spectra}

{The first example simulates spectra from a mixture of two compounds that result in two orthogonal components. The simulation has the following generating rules}:

\begin{itemize}
	\item[a)]  The first component has non-zero values for variables \{1-10\} and the second for variables \{11-20\}. For each component, loadings follow a shape of spectra: \{0.1 0.3 0.5 0.7 0.9 0.9 0.7 0.5 0.3 0.1\}. Note loadings are intentionally {non-negative}, orthogonal and sparse (they contain 0's).
	\item[b)] The scores for five individuals are generated following the pattern: 
	
		
		\begin{equation*}
		\mathbf{T}(5 \times 2) = 
		\begin{bmatrix}
					0.5& 0& 0.5& 0& 0.5 \\
					0& 0.25& 0& 0.25& 0 \\	
			\end{bmatrix}^\tT \\
		\end{equation*}
		
Note that the scores are intentionally  {non-negative} and orthogonal.
	
	
	\item[c)] The data set $\mathbf{X}$ is generated with dimension $5 \times 20$ following {$\mathbf{X} = \mathbf{T}\mathbf{P}^\tT$}. There is no noise in the data, which means that the data is purely rank 2 and the true profiles are the ones shown above. {The low dimension in observations and variables and the absence of noise are not realistic, but are useful to illustrate the performance of the models in a simple scenario.}
\end{itemize}

\begin{figure}
	\centering
	\subfigure{\includegraphics[width=0.3\textwidth]{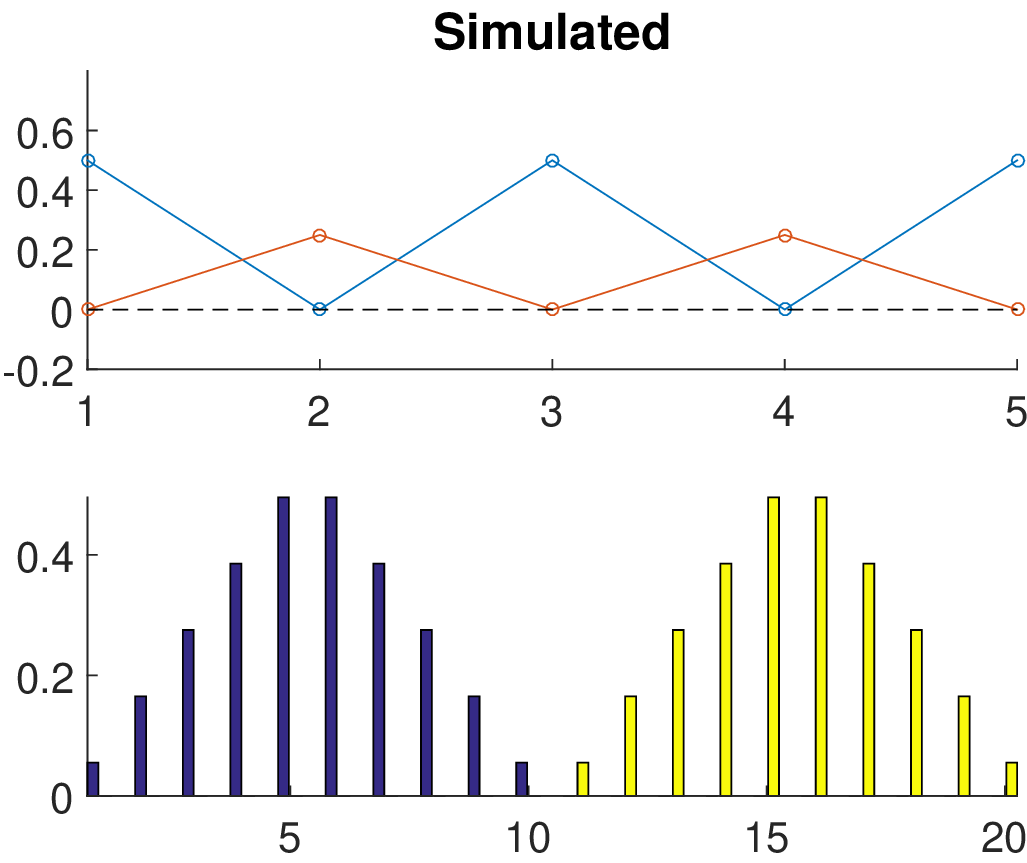}}\\
	\subfigure{\includegraphics[width=0.3\textwidth]{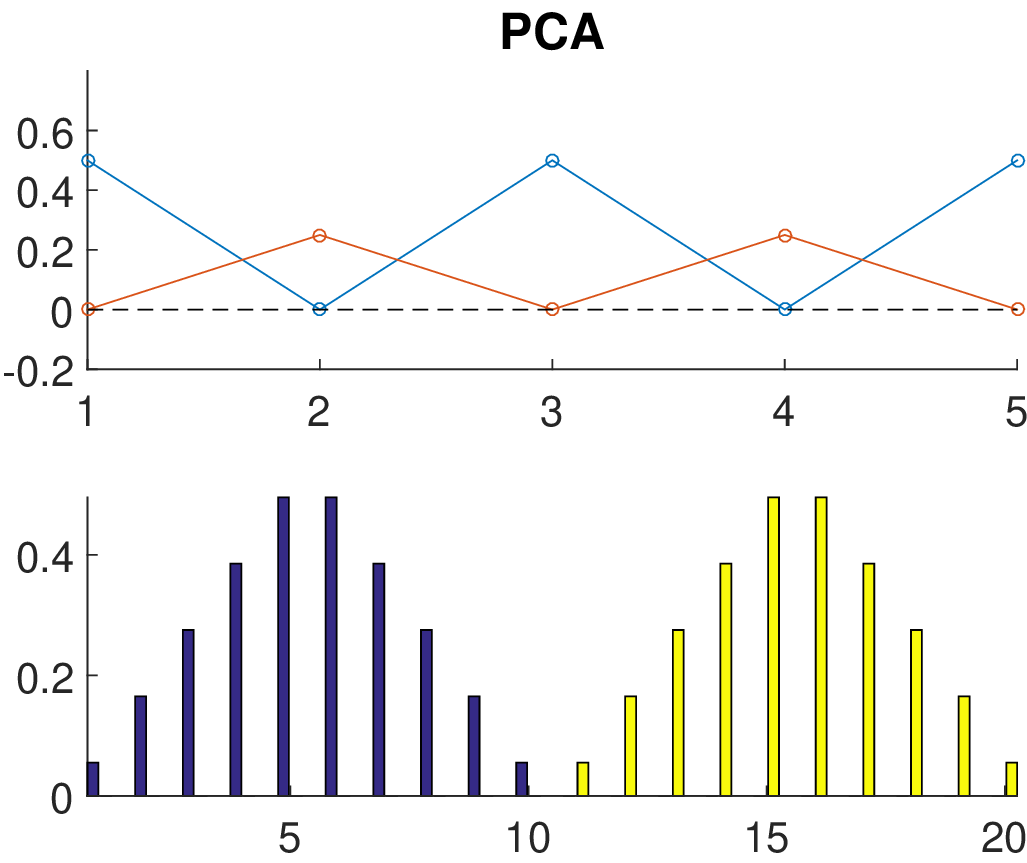}} 
	\subfigure{\includegraphics[width=0.3\textwidth]{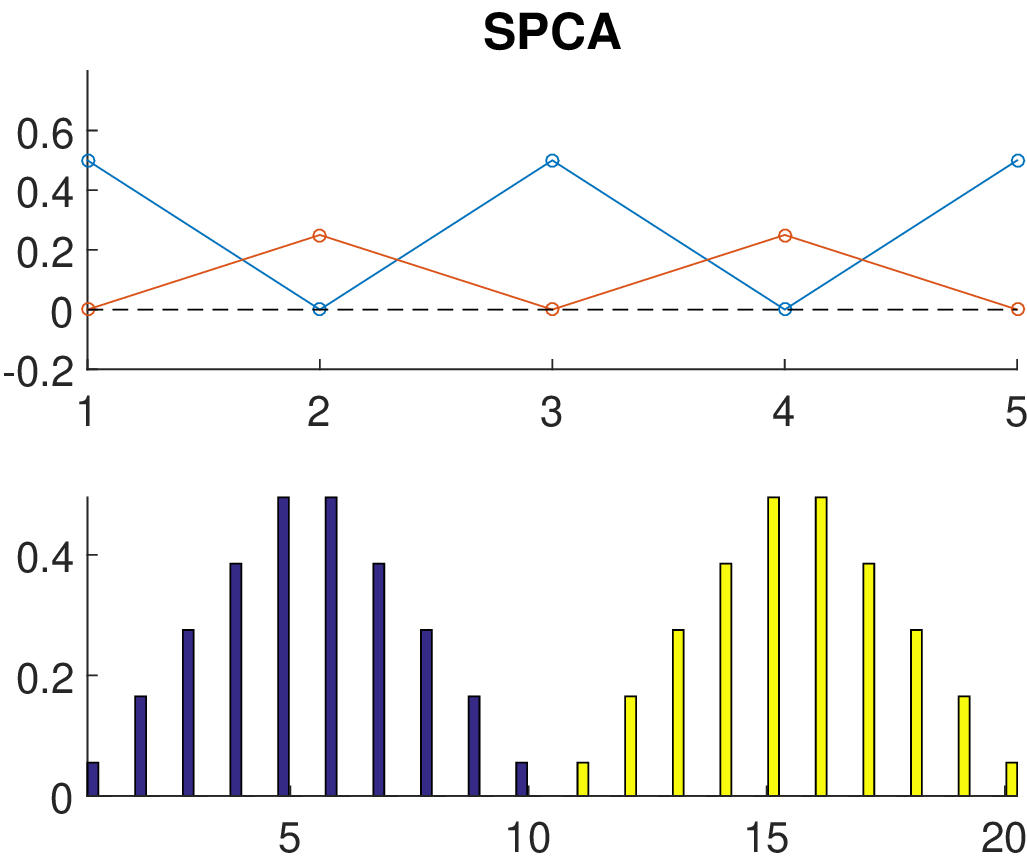}}
	\subfigure{\includegraphics[width=0.3\textwidth]{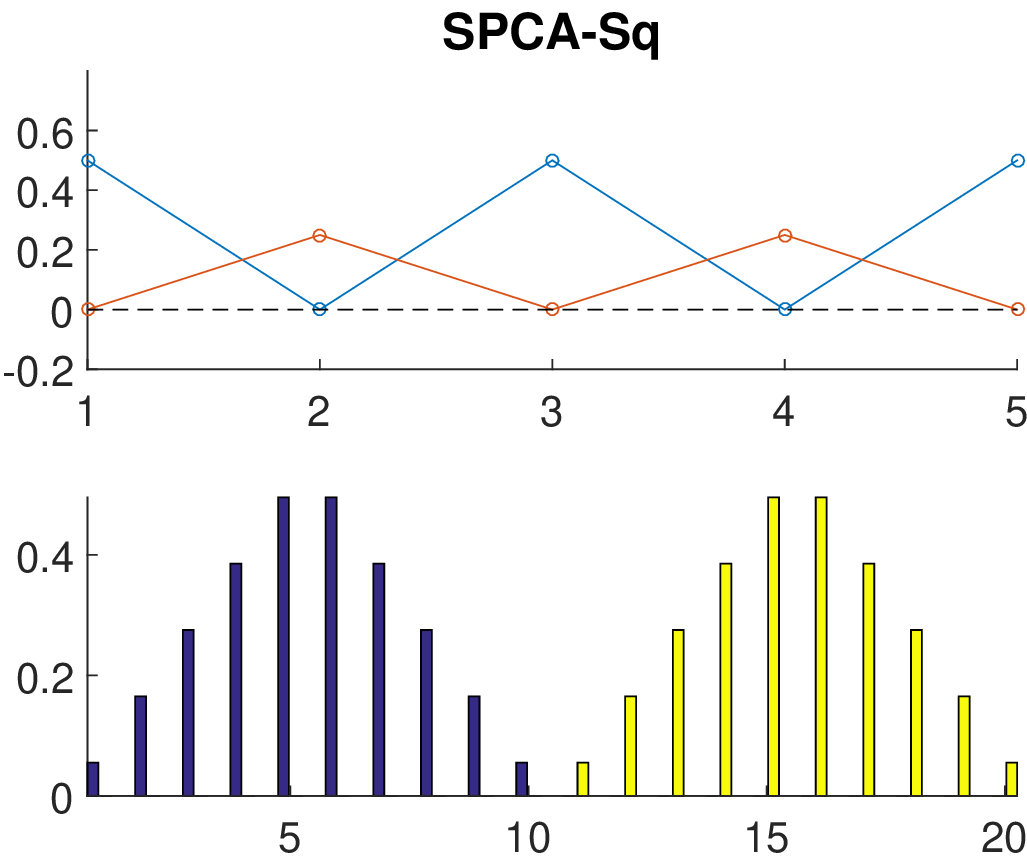}}
	\subfigure{\includegraphics[width=0.3\textwidth]{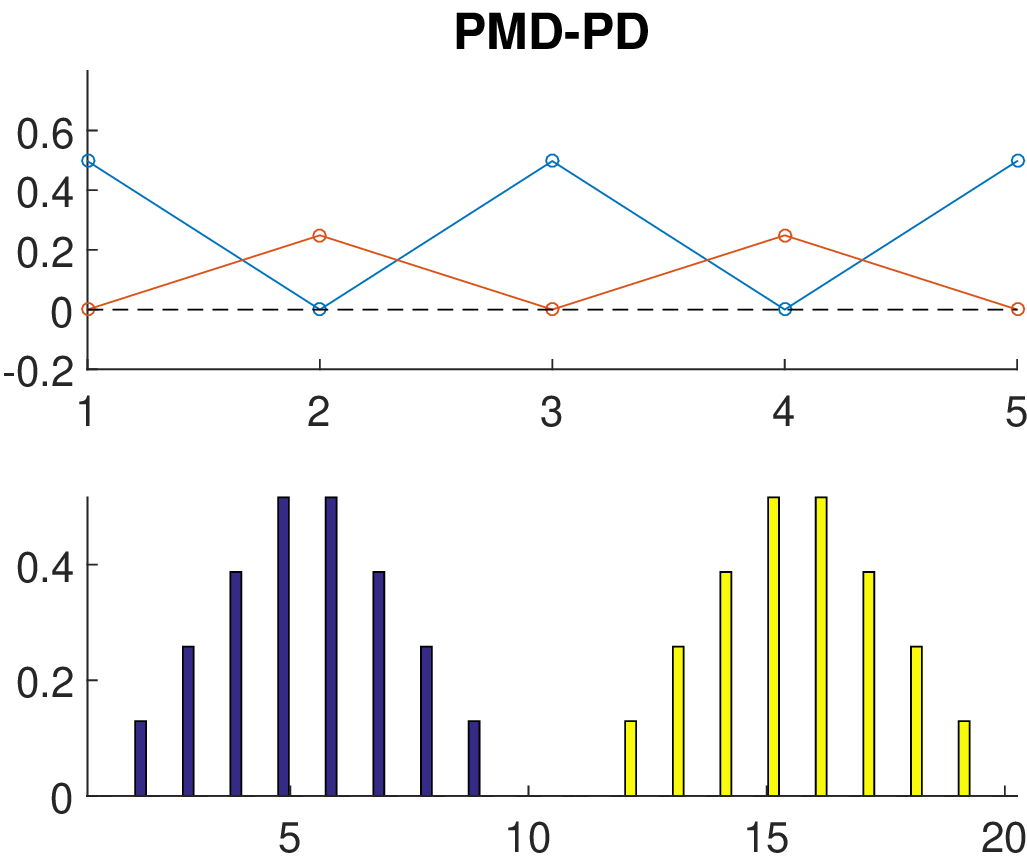}}
	\subfigure{\includegraphics[width=0.3\textwidth]{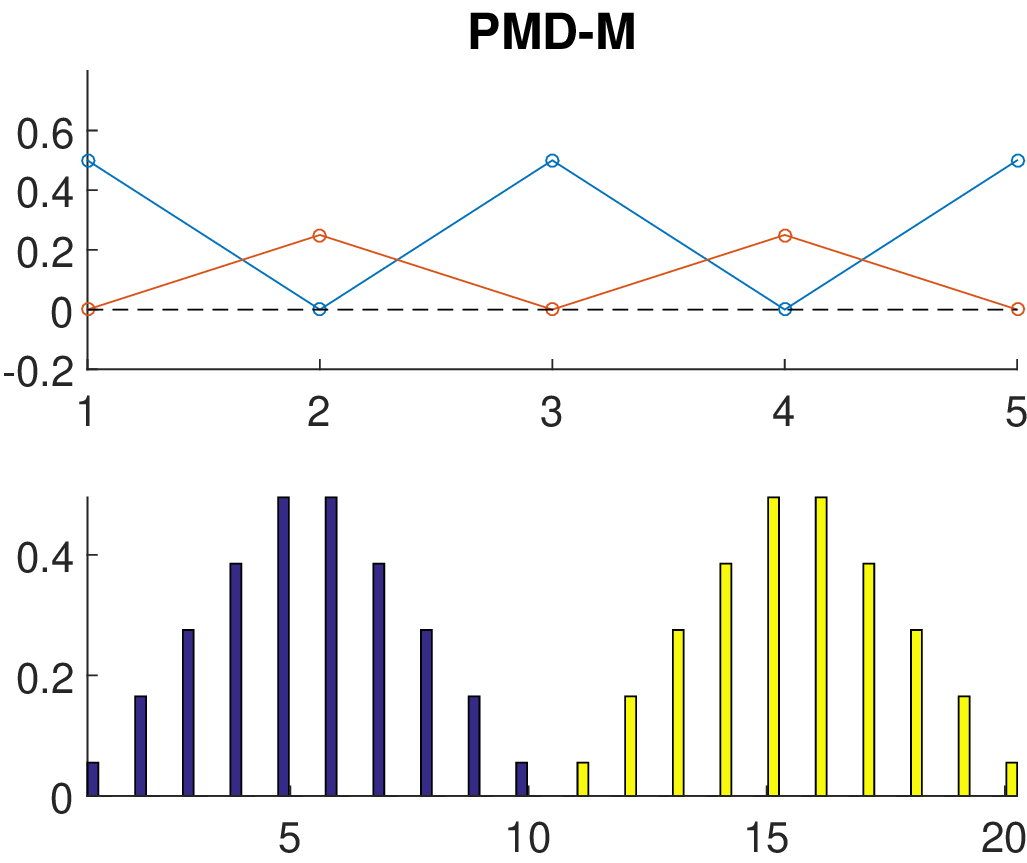}}
	\subfigure{\includegraphics[width=0.3\textwidth]{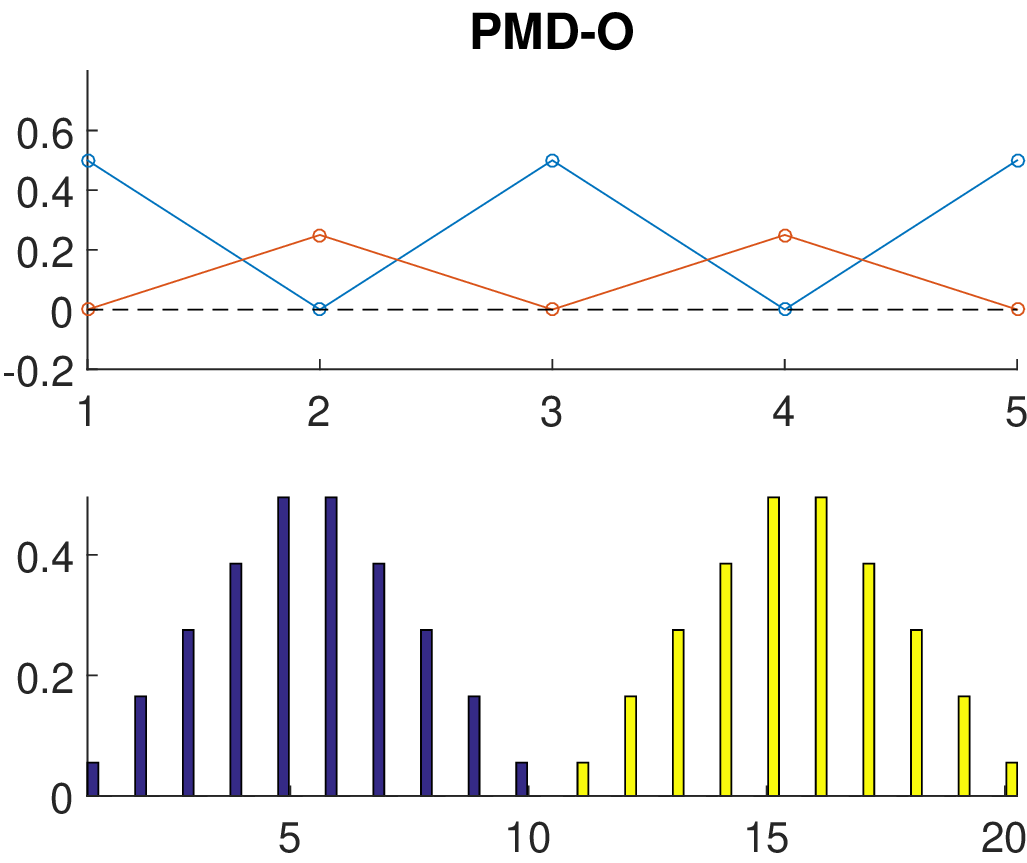}}
	\subfigure{\includegraphics[width=0.3\textwidth]{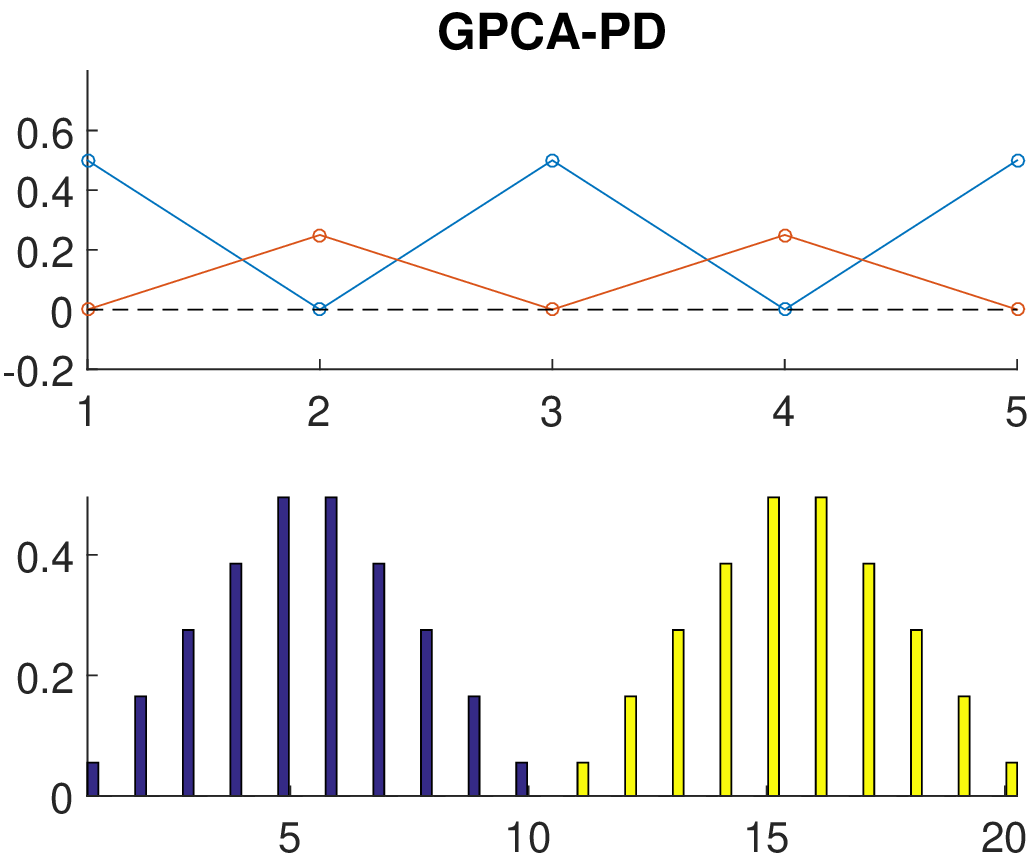}}
	\subfigure{\includegraphics[width=0.3\textwidth]{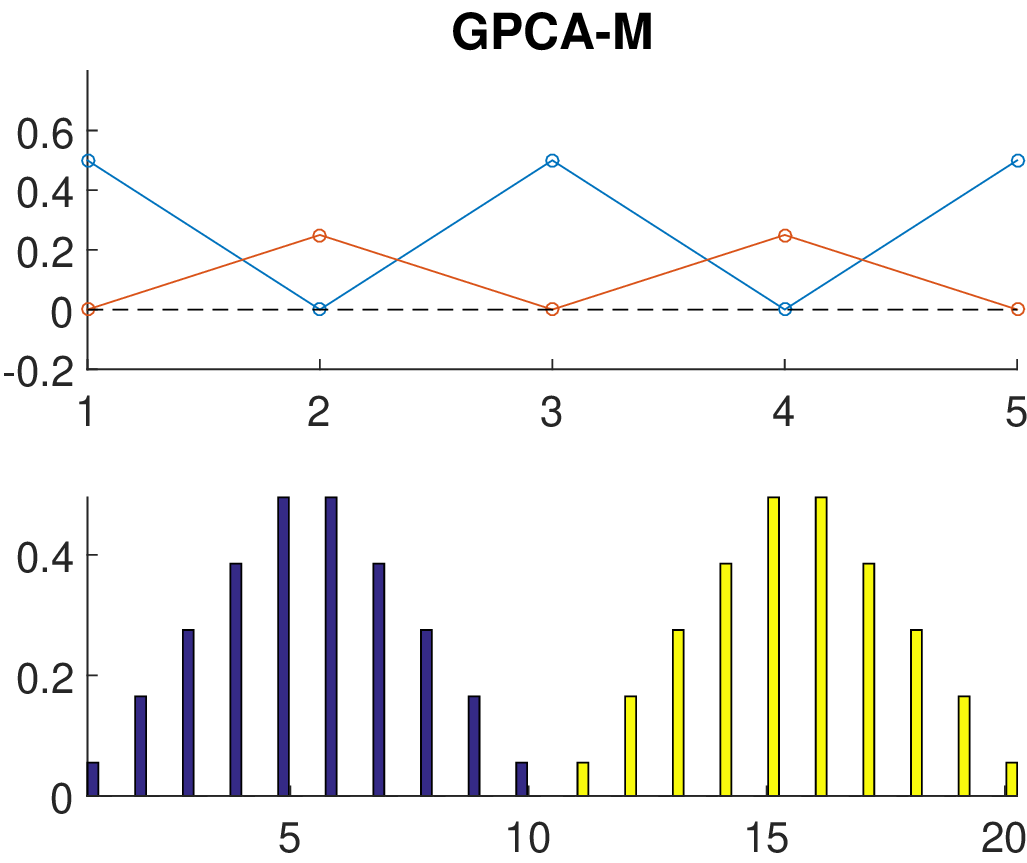}}
	\subfigure{\includegraphics[width=0.3\textwidth]{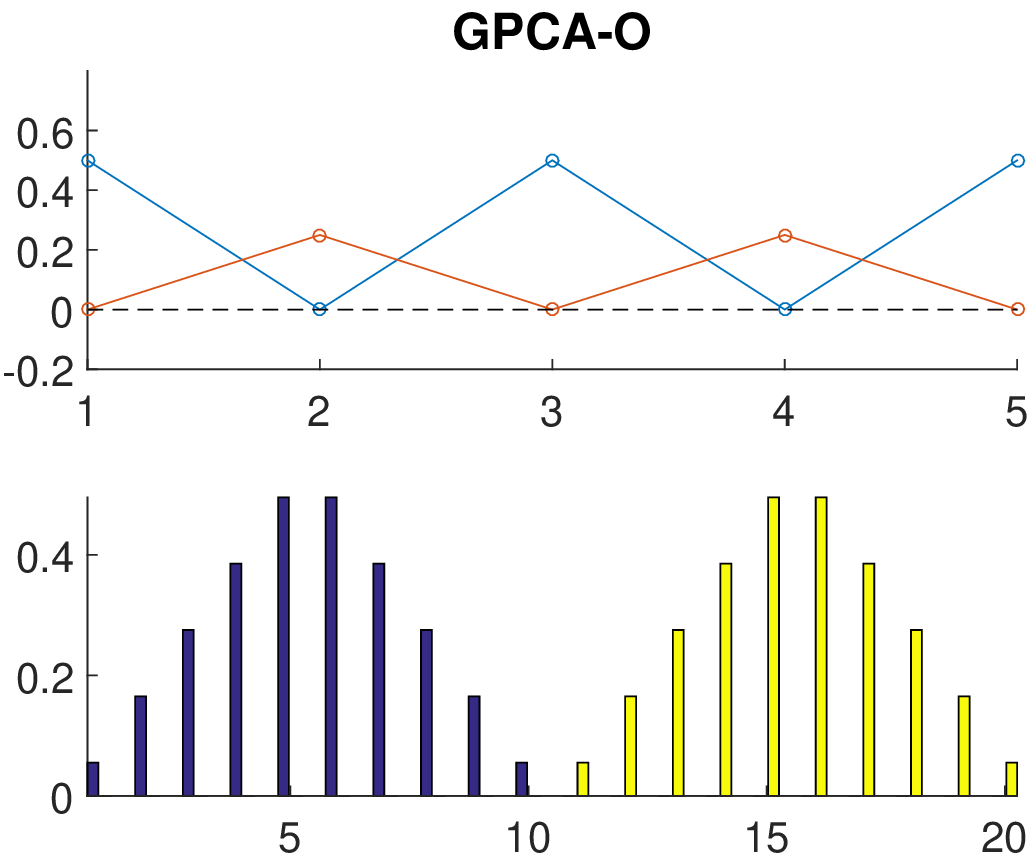}}
	\caption{Orthogonal spectra example: true data and matrix factorization methods.}
	\label{fig:Figure_ej0}
\end{figure}

The data set $\mathbf{X}$ is decomposed using PCA and the different variants of sPCA, and results are shown in Figure \ref{fig:Figure_ej0}. Data is not mean-centered and components are orthogonal}. Since all methods perfectly capture the data structure, and are equally interpretable, we shall conclude that they have equal performance when applied to noise-free, orthogonal data, without the {requirement} to impose orthogonality.

\subsection{Non-orthogonal spectra}

In the second example, we generate data from a set of sparse components that simulate non-orthogonal spectra with the following generating rules:

\begin{itemize}
	\item[a)]  Three components $\mathbf{P}$($20 \times 3$) are simulated: the first has non-zero values for variables \{1-10\}, the second for variables \{6-15\} and the third for \{11-20\}. For each component, the loadings follow a spectral shape like in the example in the previous section. Note loadings are intentionally {non-negative and} overlapping, and therefore are non-orthogonal, and sparse.
	\item[b)] The scores for five individuals are generated following the pattern:
%
			\begin{equation*}
	\mathbf{T}(5 \times 3) = 
	\begin{bmatrix}
	0.5& 0.5& 0.5& 0.5& 0 \\
	0.25& 0& 0.25& 0& 0.25 \\
	0& 0.125& 0& 0.125& 0 \\	
	\end{bmatrix}^\tT \\
	\end{equation*}

	Note that the scores for the first component are not orthogonal to the other two, but they are all non-negative.
	%
	%
	\item[c)] The data set $\mathbf{X}$ is generated with dimension $5 \times 20$ following {$\mathbf{X} = \mathbf{T}\mathbf{P}^\tT$}. No noise is introduced in the data, which means that the data is purely rank 3.
\end{itemize}
\begin{figure}
	\centering
	\subfigure{\includegraphics[width=0.3\textwidth]{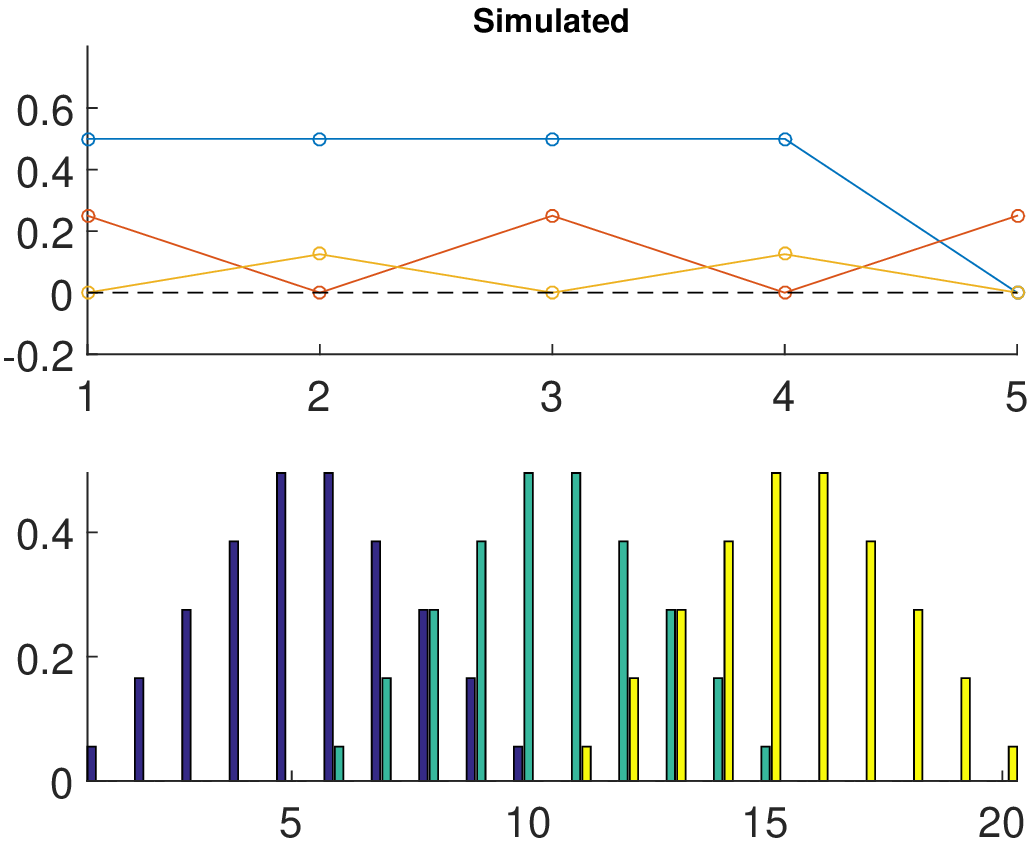}}\\
	\subfigure{\includegraphics[width=0.3\textwidth]{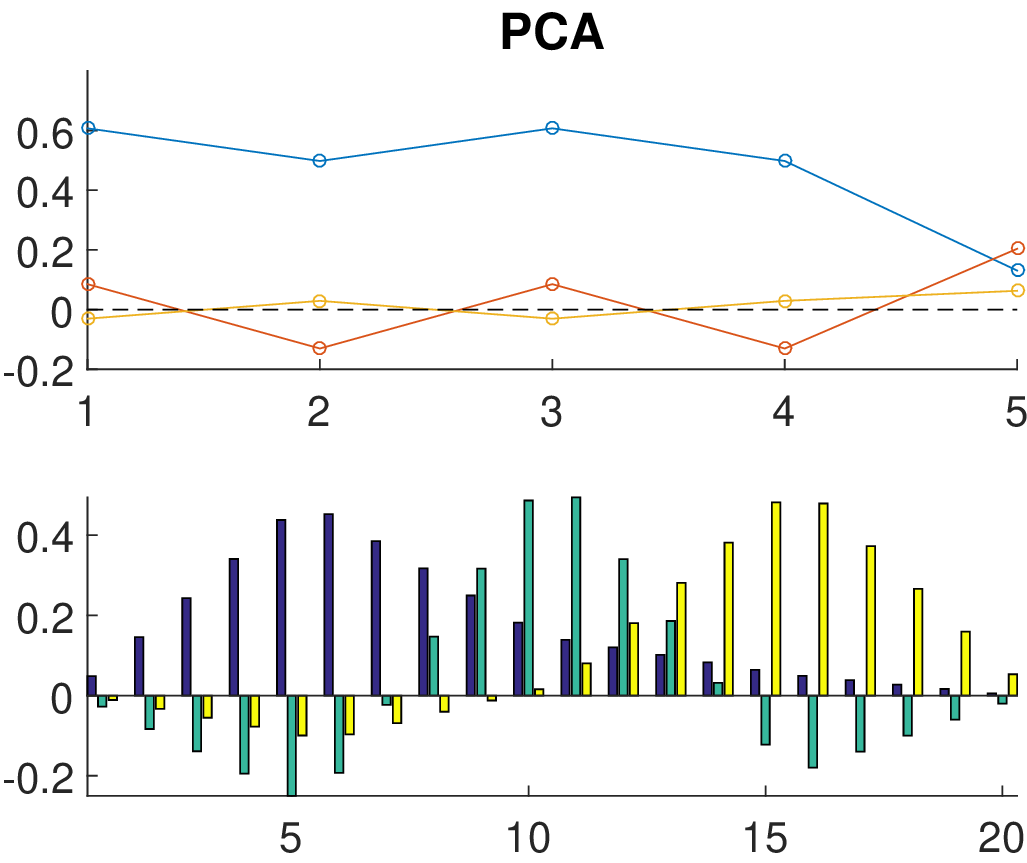}} 
	\subfigure{\includegraphics[width=0.3\textwidth]{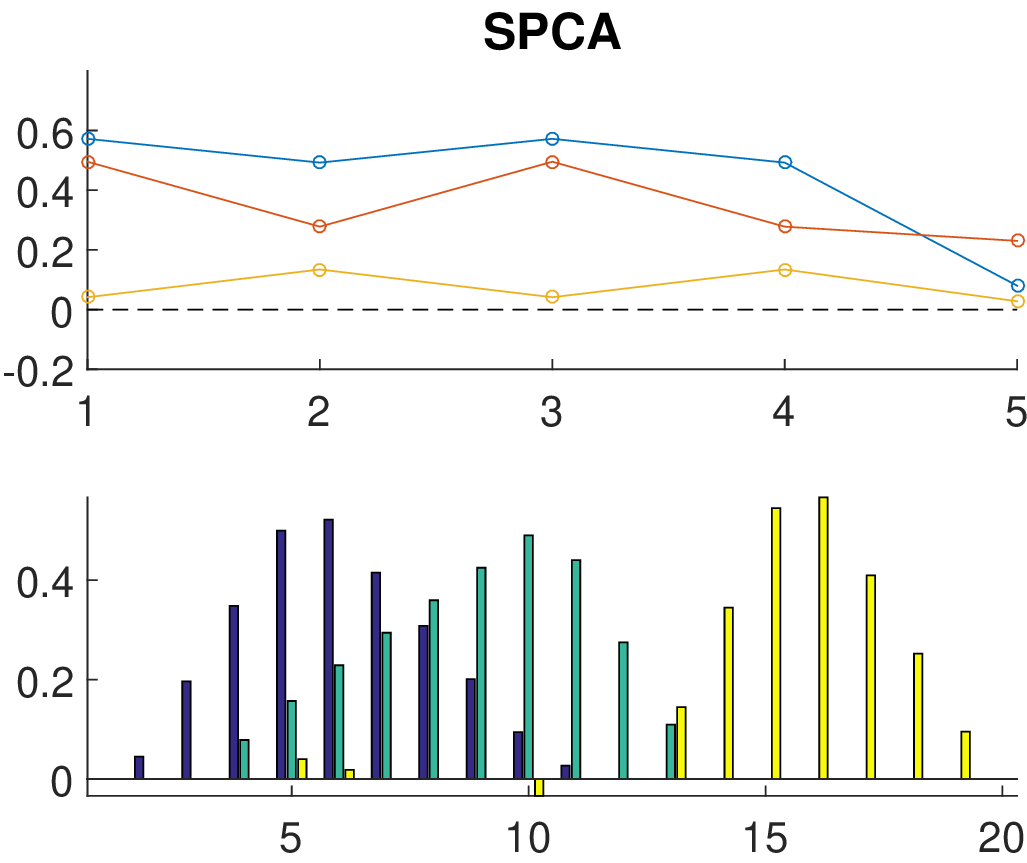}}
	\subfigure{\includegraphics[width=0.3\textwidth]{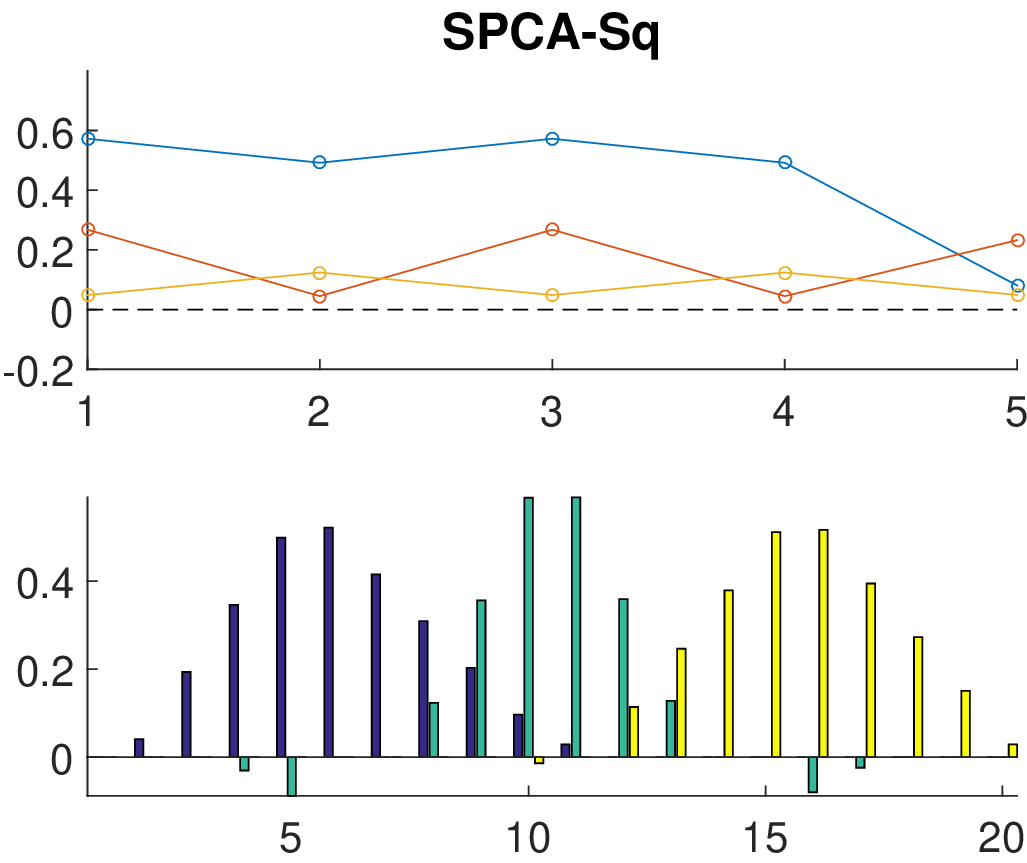}}
	\subfigure{\includegraphics[width=0.3\textwidth]{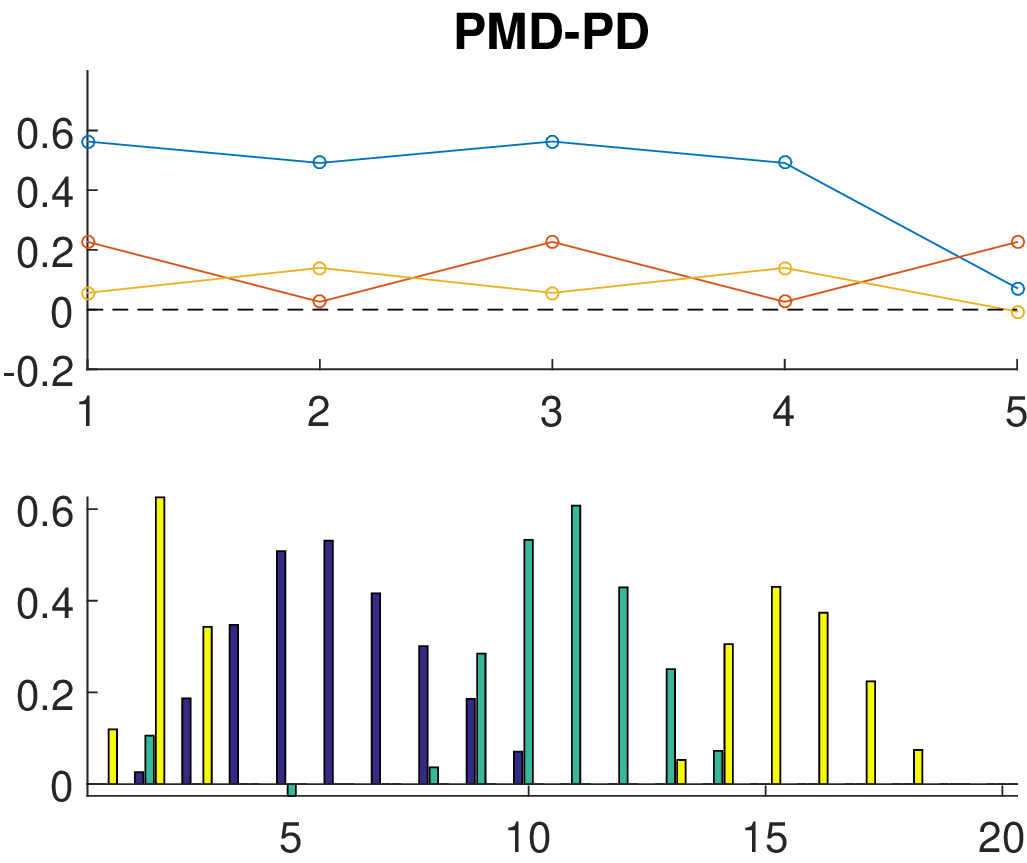}}
	\subfigure{\includegraphics[width=0.3\textwidth]{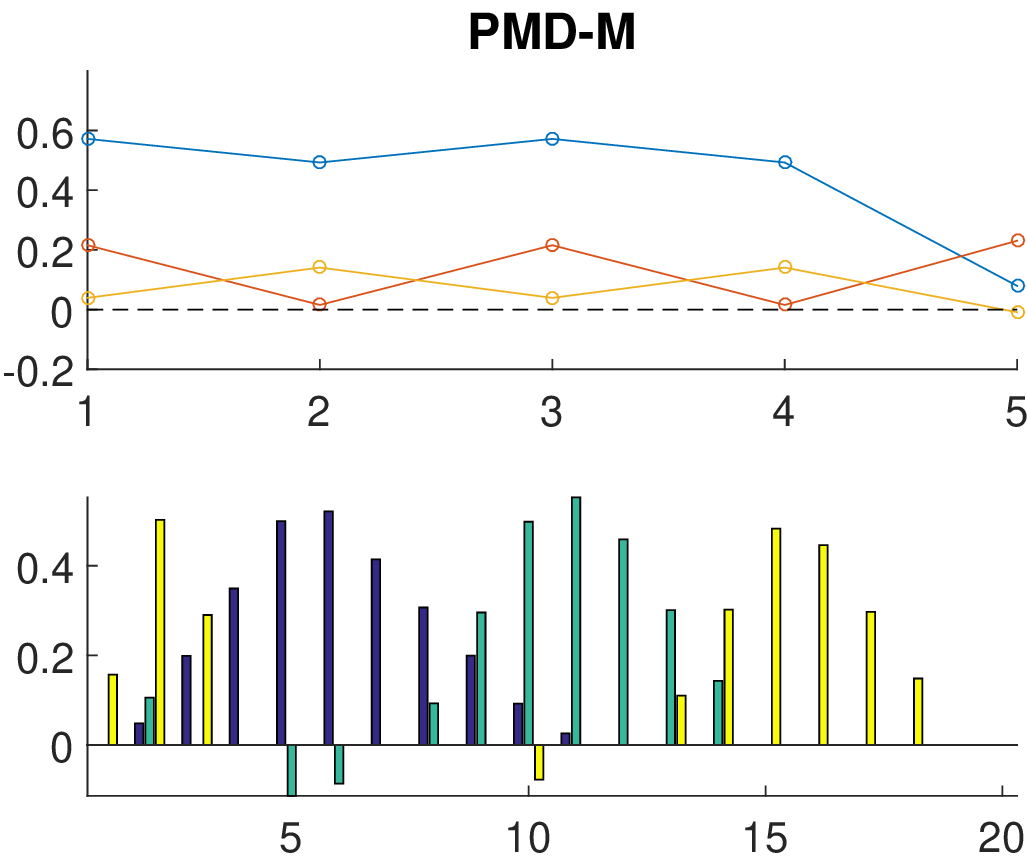}}
	\subfigure{\includegraphics[width=0.3\textwidth]{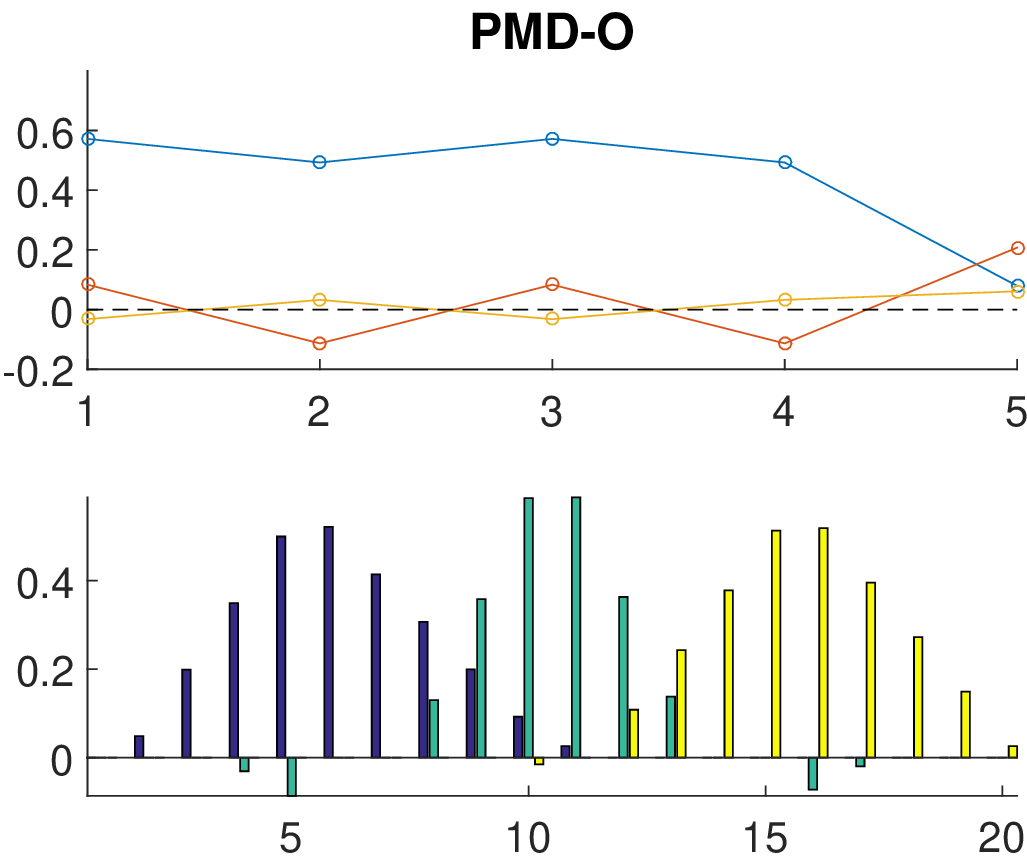}}
	\subfigure{\includegraphics[width=0.3\textwidth]{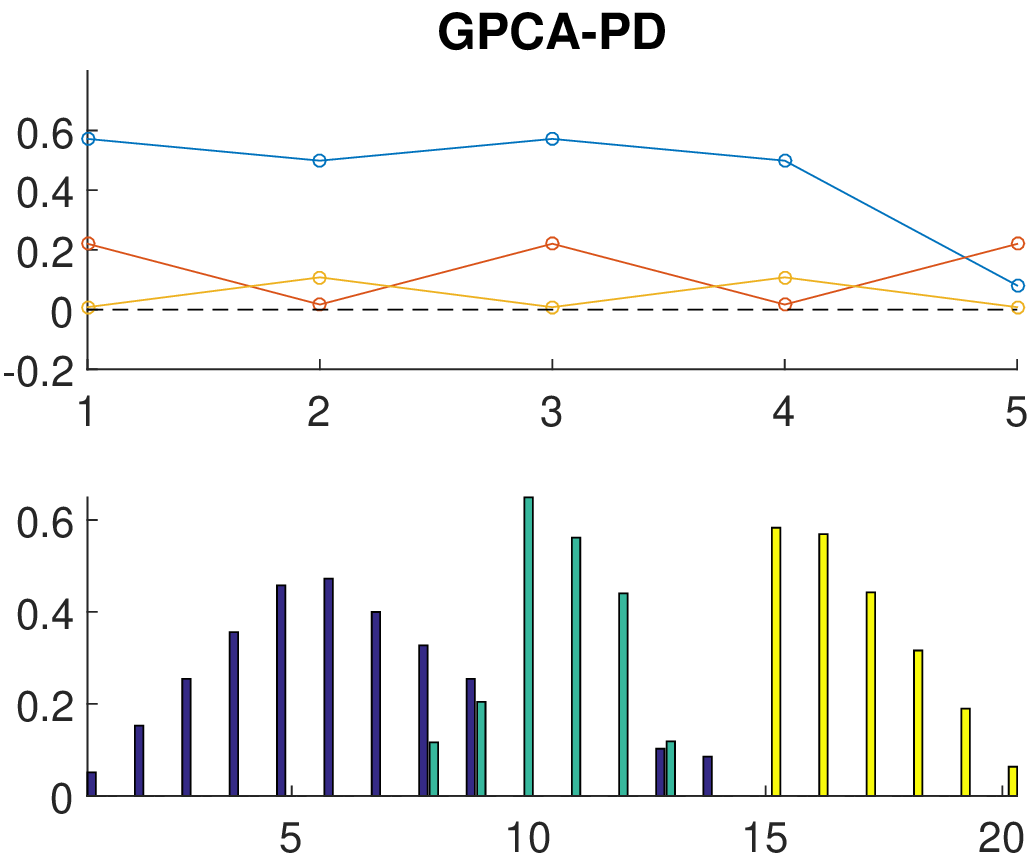}}
	\subfigure{\includegraphics[width=0.3\textwidth]{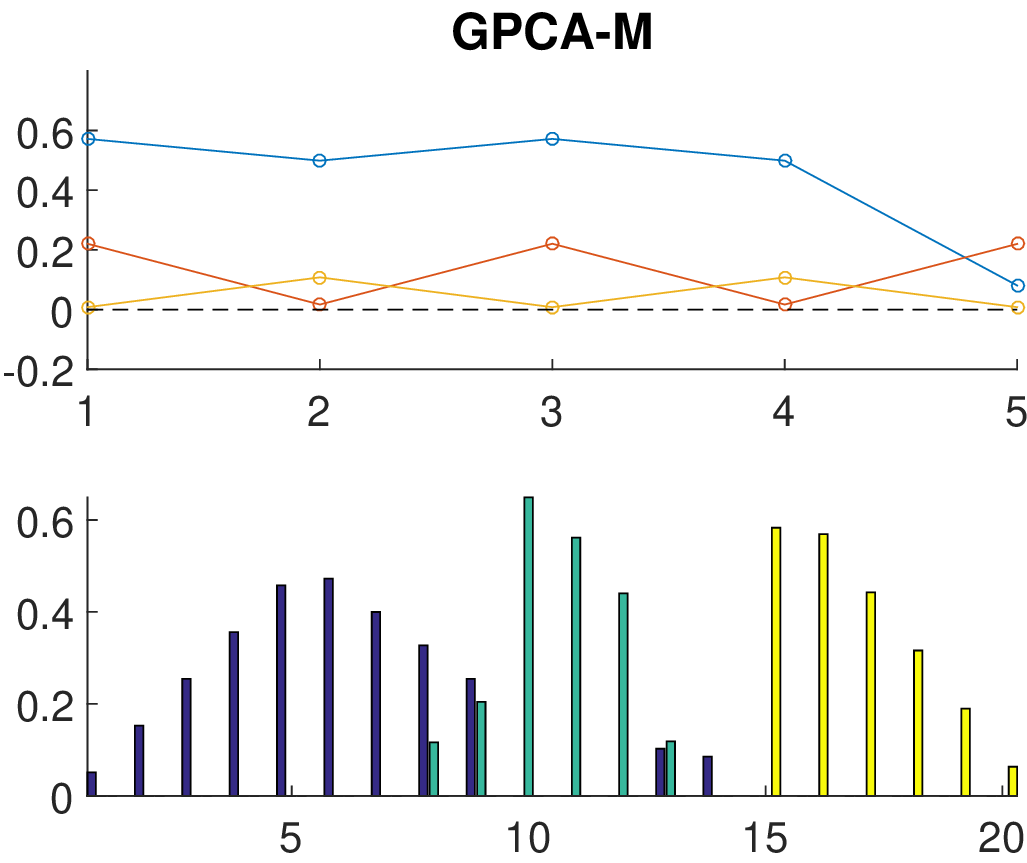}}
	\subfigure{\includegraphics[width=0.3\textwidth]{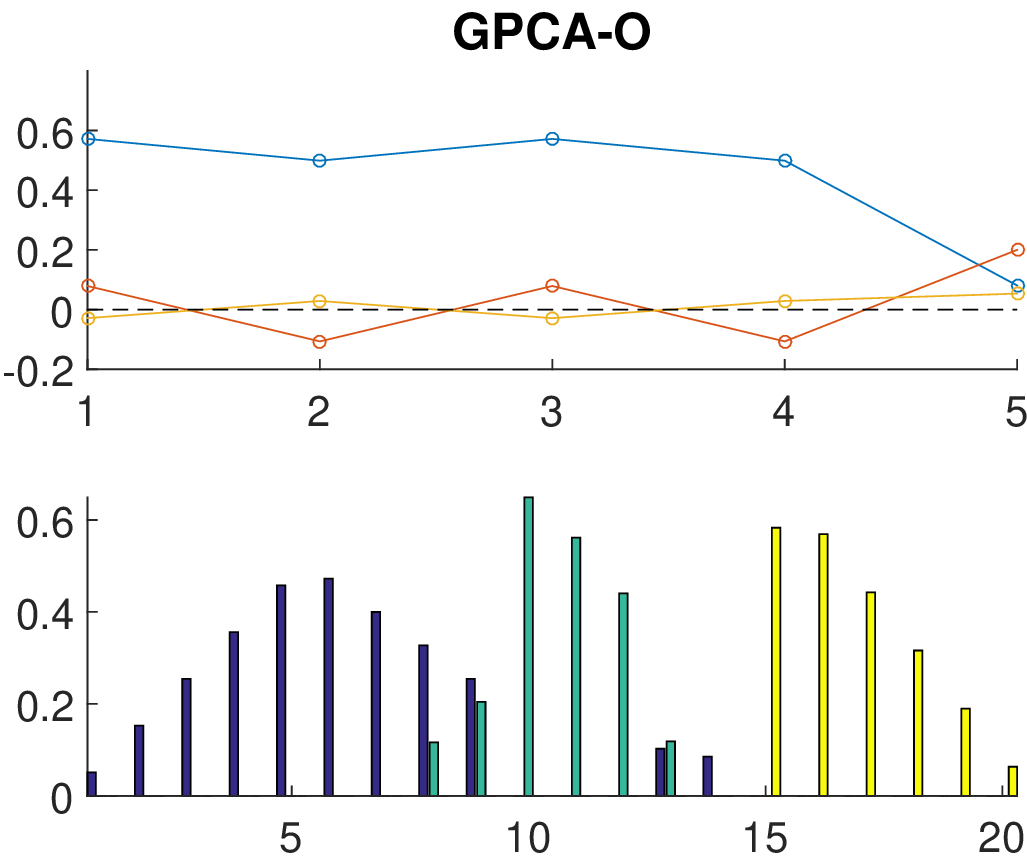}}
	\caption{Non-orthogonal spectra example: true data and matrix factorization methods.}
	\label{fig:Figure_ej1}
\end{figure}
Figure \ref{fig:Figure_ej1} shows the result of the simulation and the approximation of data $\mathbf{X}$ by the sparse factorization methods considered along with PCA. Scores were intentionally computed following the standard approach as $\mathbf{\hat{T}} = \mathbf{X}\mathbf{\hat{P}}$, with $\mathbf{\hat{P}}$ the loadings {estimated} in the algorithms. PCA shows two limitations and fails to reproduce the structure of the data: first, components are not sparse and are constrained to be orthogonal and this induces distortion in the reconstructed spectra; second, scores must also be orthogonal, and to satisfy this condition some score values have to be negative to compensate for the positive ones. 
Concerning sparse models, they show a disparate performance. The sequential version of SPCA visually outperforms the original SPCA and most other methods in the approximation. GPCA-PD and GPCA-M are also very accurate, but have difficulties to model {the second and third} loading vector. PMD is also less accurate in determining the loadings. Regarding the scores, orthogonalized methods distort the simulated (non-orthogonal) scores, as expected, and give the same score patterns as PCA. 

\subsection{Monte Carlo simulation}

Based on the previous examples, we designed a Monte Carlo study in which datasets are generated where the components can have a random number of non-zero loadings and the values of the scores and loadings are also randomly selected. With respect to the previous illustrative examples, here the data sets are simulated  1) with more realistic dimensions and structure ($50\times200$ and 5 components) as encountered in practical applications, and 2) are not constrained to be non-negative. Under this simulation scheme, the scores and the loadings are likely correlated, but at a lesser extent than in the case of non-orthogonal spectra in previous section. The generating rules are:  

\begin{enumerate}
	
	\item {Five components $\mathbf{P}$ ($200 \times 5$) are simulated. 
	For each simulated data the loading matrix $\bP \sim \mathit{N}(0,1)$ has size $200 \times 5$, for a total of 1000 loadings. An auxiliary $200 \times 5$ matrix $\bW \sim \mathit{N}(0,1)$ is generated, so that for elements of $\bW$ below 1 the corresponding elements in $\bP$ are set to 0 ($W_{ij} < 1 \implies P_{ij} = 0$). This procedure generates, on average, 160 non-zero loadings for every 1000 loadings simulated. The resulting loading vectors have a different degree of sparsity, contain both positive and negative values and are usually correlated because the non zero elements can overlap.}
	%
	%
	%
	\item The scores for 50 observations are randomly generated {from $\mathit{N}(0,1)$} {and are multiplied by a constant so that their variance follows an exponential decay, similar to  previous examples. The scores can be non-orthogonal and contain either positive or negative values.}
	%
	%
	\item The data set $\mathbf{X}$ is generated with dimension $50 \times 200$ but no noise is added, like in the previous examples. This means that the data is purely rank 5.
\end{enumerate}

{The simulation is repeated {100} times. All sparse algorithms considered require the setting of one or more metaparameters. We optimistically selected the metaparameters for each method to meet on average the known level of sparseness of the data generated. In this way, all methods can be fairly compared.} 
%

%
%

Results from the non-orthogonal spectra and Monte Carlo simulation will be used to support the discussion in the following sections.

\section{Correlation of Scores and Loadings}

As previously discussed, scores and loadings in sPCA can be correlated. To quantify to which extent this happens, we defined and computed the following statistics 
{
\begin{itemize}
	\item MACS: Mean absolute correlation coefficient of the scores, measured as the mean of the absolute value of the elements below the diagonal of $\mathbf{\hat{T}}^\tT\mathbf{\hat{T}} \oslash \|\diag(\mathbf{\hat{T}}^\tT\mathbf{\hat{T}})\|_2$, where $\oslash$ is the Hadamard (element-wise) division. 
	This parameter ranges between 0 (absence of correlation) and 1 (total correlation). 
	
	\item MACL: Mean absolute correlation coefficient of the loadings, measured as the mean of the absolute value of the elements below the diagonal of $\mathbf{\hat{P}}^\tT\mathbf{\hat{P}}$, for normalized loadings. This parameter ranges between 0 (absence of correlation) and 1 (total correlation). 
\end{itemize}
}

Results are shown in Table \ref{table:statsa_Corr} for the non-orthogonal spectra and in {Figures 
\ref{fig:Figure_ej100}(a) and \ref{fig:Figure_ej100}(b)} for the Monte Carlo simulation. The actual correlation of the simulated data is shown as a baseline. It can be seen that all but the orthogonalized versions provide correlated scores, and that all methods produce correlated loadings. {Non-orthogonal methods seem to largely overestimate the true correlation in the scores, while orthogonal methods obviously underestimate it {(they yield exactly 0 correlation in all cases)}. The sequential approaches tend to underestimate the correlation in loadings}. 


\begin{table}
	\centering
	\footnotesize
	\begin{tabular}{|c|c|c|c|c|c|c|c|c|c|}
		\hline
		 & \textbf{Simulated}
		 & \textbf{sPCA} & \textbf{sPCA-Sq} & \textbf{PMD-PD} & \textbf{PMD-O} & \textbf{PMD-M} & \textbf{GPCA-PD} & \textbf{GPCA-M} & \textbf{GPCA-O} \\
		\hline           
\textbf{MACS} & 0.43 & 0.84 & 0.73 & 0.66 & 0 & 0.58 & 0.52 & 0.52 & 0 \\                                                         
\hline                                                                                                                                           
\textbf{MACL} & 0.17 & 0.21 & 0.05 & 0.09 & 0.05 & 0.07 & 0.03 & 0.03 & 0.03 \\ 
		\hline
	\end{tabular}
	\caption{{Correlation in scores and loadings for the non-orthogonal spectral example. MACS: Mean absolute correlation coefficient of the scores; MACL: Mean absolute correlation coefficient of the loadings}}
	\label{table:statsa_Corr}
\end{table}

	\begin{figure}
	\centering
	\subfigure[]{\includegraphics[width=0.32\textwidth]{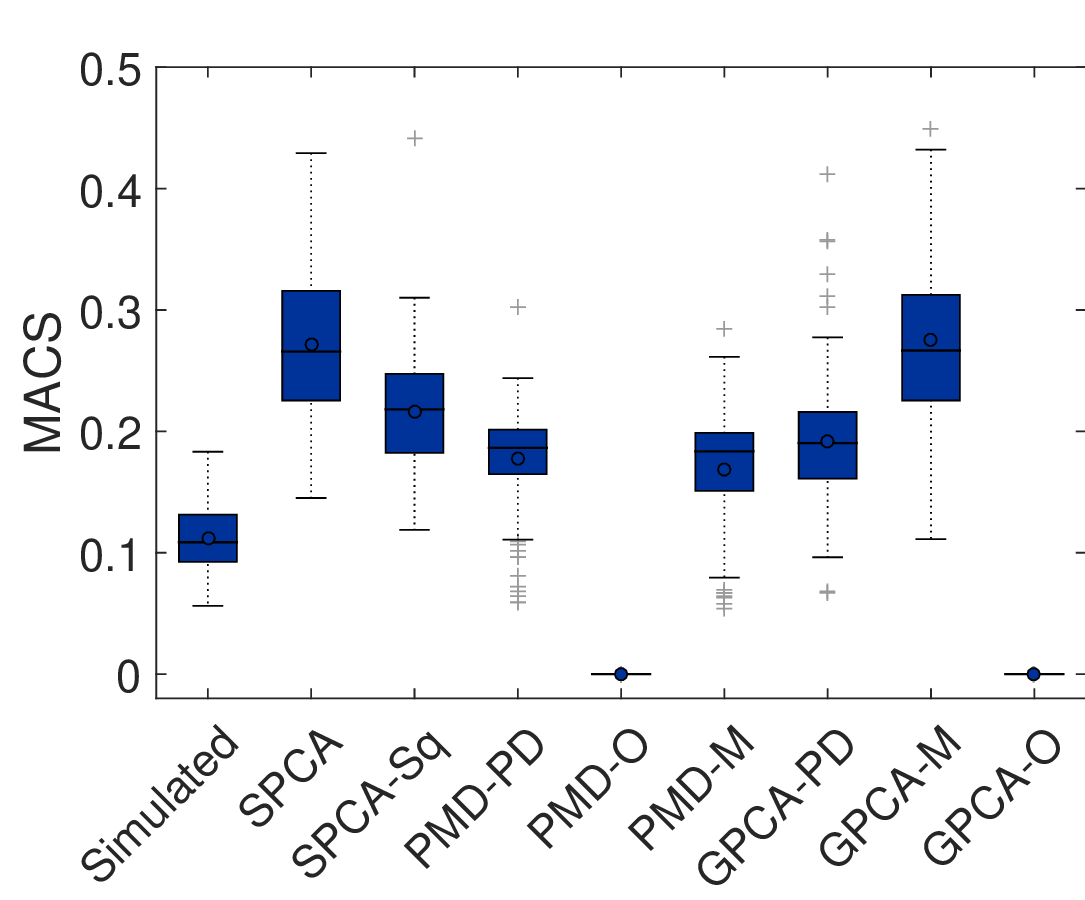}}
	\subfigure[]{\includegraphics[width=0.32\textwidth]{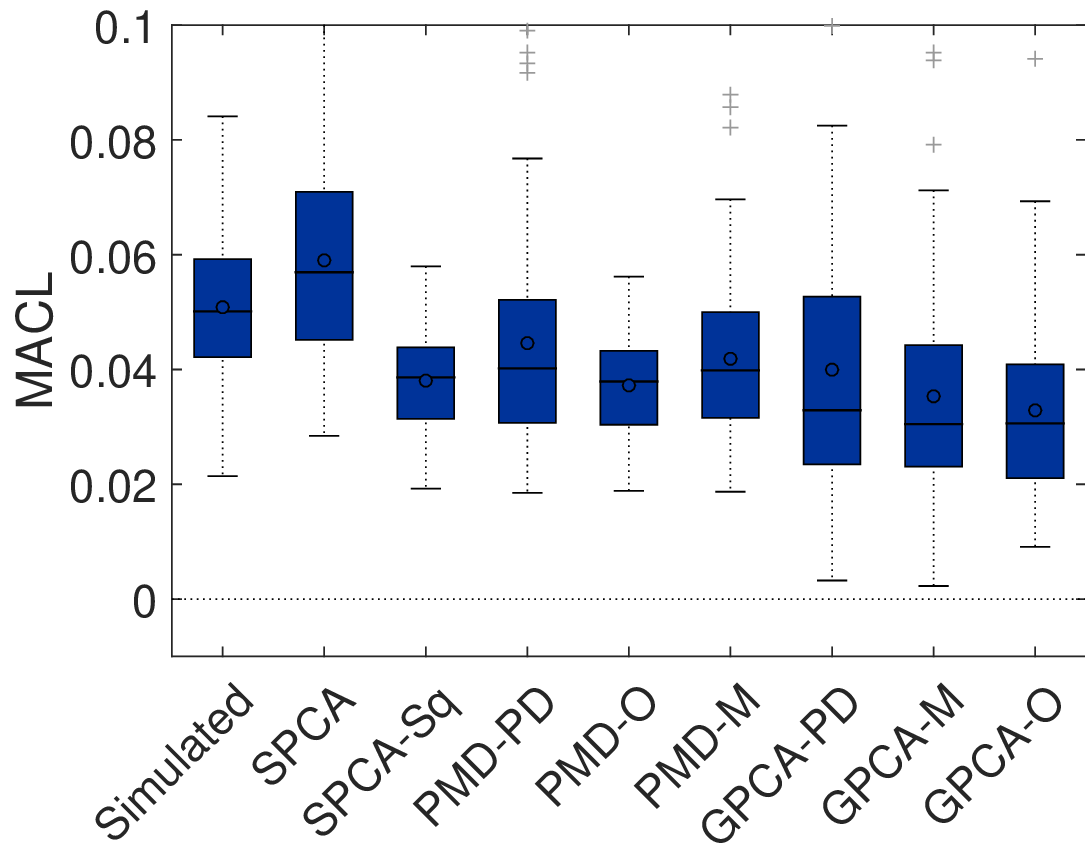}}
	\subfigure[]{\includegraphics[width=0.32\textwidth]{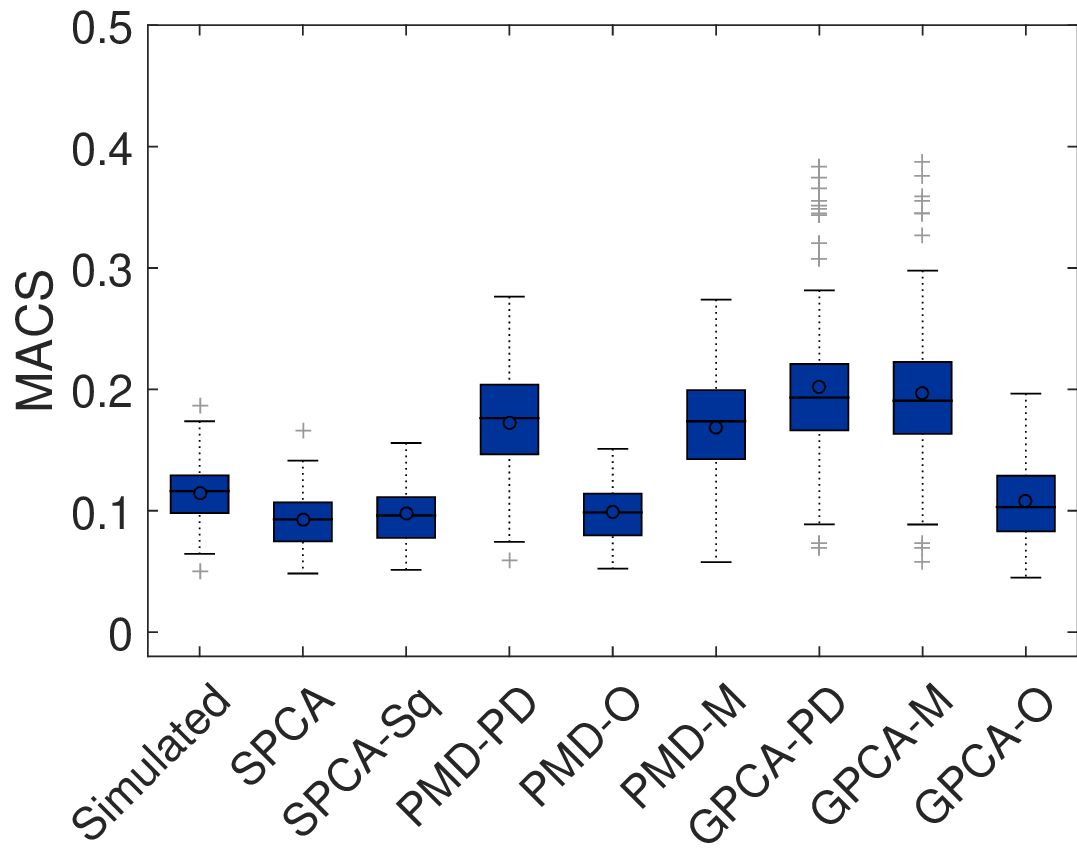}}

	\caption{Box Plots of MACS (a), MACL (b) and MACS with corrected scores (c) for the Monte Carlo simulation. Please, note y-axis are of different scale.}
	\label{fig:Figure_ej100}
\end{figure}

From these results, we can infer that correlation in scores and loadings is commonplace. While this can be beneficial for the interpretation of complex data that can be, in general, non-orthogonal, it also poses several problems and limitations that should be taken into account when analyzing a sparse model.
{Correlated loadings distort distances in the scatter plots \cite{Kiers,Geladi}, while {correlated scores and/or loadings may lead to the incorrect estimation of explained and residual variance \cite{Mackey2008}.}
More implications resulting from the correlation of loadings and scores will be described in the next sections and in the follow-up paper.}


%

\section{Computation of Scores, Residuals and Variance}
{The non-orthogonality of the loadings has critical implications for the computation of the scores, the residuals and explained variance. 
While in standard PCA the scores are calculated as 
\begin{align}\label{PCAscores}
	\mathbf{\hat{T}} = \mathbf{X} \mathbf{\hat{P}}.
\end{align}
in the sparse PCA setting the loadings should be computed using the correction formula
\begin{equation} \label{scores}
\mathbf{\hat{T}} = \mathbf{X} \mathbf{\hat{P}} (\mathbf{\hat{P}}^\tT \mathbf{\hat{P}})^+  
\end{equation}
\noindent since Eq. (\ref{PCAscores}) cannot be applied because loadings are, in general, not orthogonal. The superscript '+' indicates the Moore-Penrose inverse.}
For orthonormal loadings, Eq. (\ref{scores}) reduces to Eq. (\ref{PCAscores}). A noteworthy characteristics of Eq. (\ref{scores}) is that the scores cannot be {accurately} computed in a sequential, one-component-at-a-time fashion, since the scores depend on the scores of the other components.}

%
%
%
%

\begin{figure}
\centering
\subfigure[scores per-component]{\includegraphics[width=0.32\textwidth]{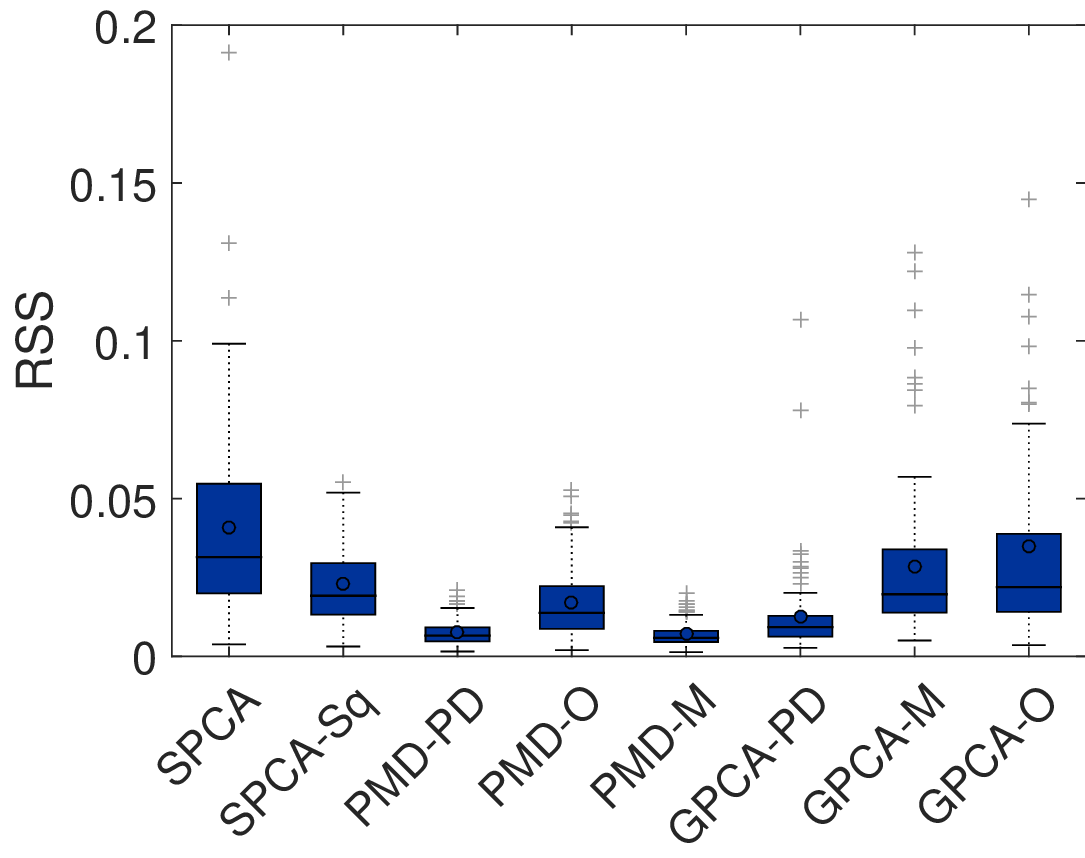}}
\subfigure[corrected scores]{\includegraphics[width=0.32\textwidth]{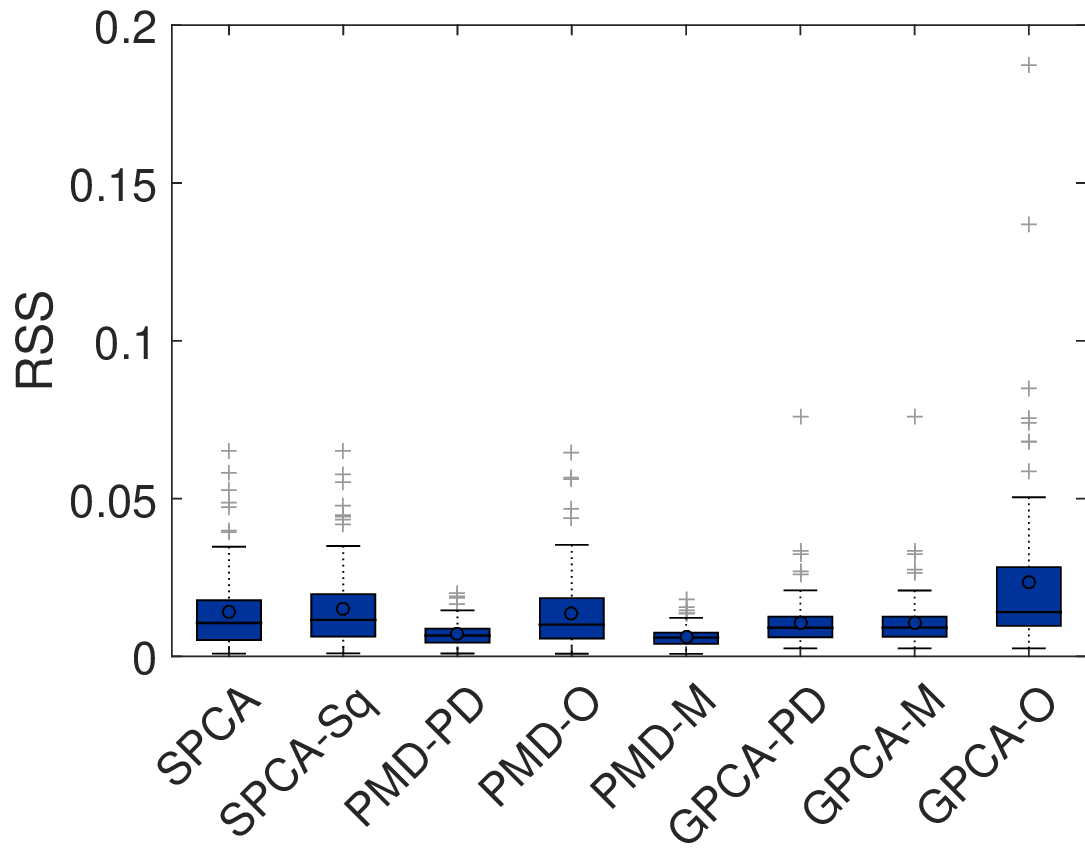}}
\caption{Box Plots of \textbf{RSS} for the Monte Carlo simulation.}
\label{fig:Figure_ej100c_RSS}
\end{figure}

{The level of inaccuracy in the computation of the residual can be assessed using a Monte Carlo simulation. Figures \ref{fig:Figure_ej100}(a) and \ref{fig:Figure_ej100}(c) show a comparison of the correlation observed in the scores computed one-at-a-time ($\mathbf{\hat{T}} = \mathbf{X}\mathbf{\hat{P}}$) and the correlation observed in the scores corrected using Eq. (\ref{scores}). It can be seen that for all methods considered, the correlation between scores is affected by the correction  and is reduced in most cases. The correction also induces correlation in orthogonalized methods.
The way correction affects the residuals can be quantified using the following RSS statistic:  
	
	\begin{itemize}
		\item RSS: The residual sum-of-squares normalized by the total sum-of-squares of the data, $\trc(\mathbf{E}^\tT\mathbf{E})$/$\trc(\mathbf{X}^\tT\mathbf{X})$, with
		$\mathbf{E} = \mathbf{X} - \mathbf{\hat{T}}\mathbf{\hat{P}}^\tT$. This quantity reflects the quality of the scores capturing the variance in the data, and should be as close to 0 as possible.
	\end{itemize}
}
{Figure \ref{fig:Figure_ej100c_RSS} compares residuals before and after correction: the correction reduces the sum-of-squares of the residuals. This reduction comes by definition, given that Eq. (\ref{scores}) is a least squares correction. Note that uncorrected residuals are used within the sPCA deflation-based algorithms to compute new scores, so this computation may be inaccurate.}   

\begin{figure}
	\centering
	\subfigure{\includegraphics[width=0.3\textwidth]{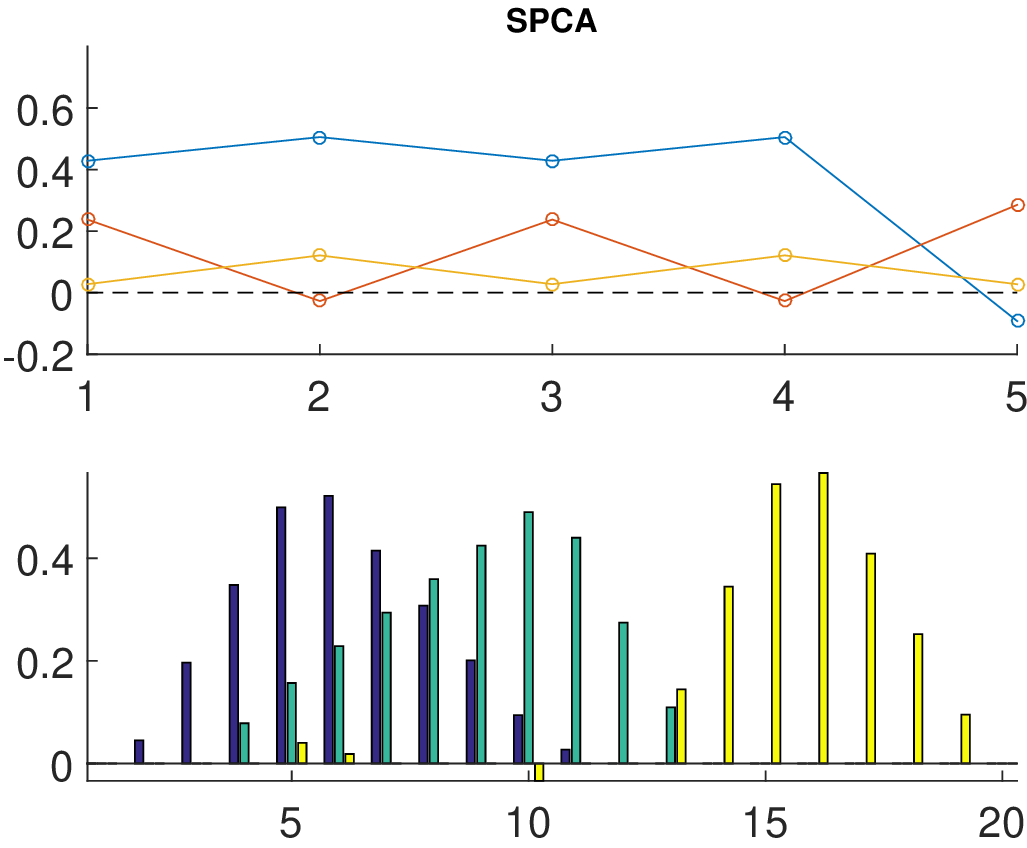}}
	\caption{Non-orthogonal spectra example: corrected sPCA.}
	\label{fig:Figure_ej1c}
\end{figure}

Figure \ref{fig:Figure_ej1c} shows the results of {correcting the scores using Eq. (\ref{scores}) in the non-orthogonal spectra example when simultaneous SPCA is used. Comparing these results with those shown in Figure \ref{fig:Figure_ej1}, it can be seen that the correction is relevant and greatly improves interpretability}. However, for sequential approaches, the correction is negligible. 

{The correct computation of scores also affects the estimation of the variance explained by each component. 
Consider a full rank approximation $\mathbf{X} = \mathbf{\hat{T}}\mathbf{\hat{P}}^\tT$. The variance of $\mathbf{X}$ can be computed as:
$\trc(\mathbf{X}^\tT\mathbf{X}) = \trc(\mathbf{\hat{P}}\mathbf{\hat{T}}^\tT\mathbf{\hat{T}}\mathbf{\hat{P}}^T)$. For orthogonal loadings, $\trc(\mathbf{X}^\tT\mathbf{X}) = \trc(\mathbf{\hat{T}}^T\mathbf{\hat{T}}\mathbf{\hat{P}}^\tT\mathbf{\hat{P}}) = \trc(\mathbf{\hat{T}}^\tT\mathbf{\hat{T}})$. However, for non-orthogonal loadings the last equality does not hold. {In general, the captured/explained variance in a component model with one sparse mode should  be measured as $\trc(\mathbf{\hat{P}}\mathbf{\hat{T}}^\tT\mathbf{\hat{T}}\mathbf{\hat{P}}^\tT)$.} A more detailed discussion can be found in the \ref{APP:Variance}.

{To support the previous discussion, we defined and calculated several statistics for (1) the scores computed as $\mathbf{\hat{T}} = \mathbf{X}\mathbf{\hat{P}}$ and (2) for the scores corrected with Eq. (\ref{scores}). These statistics are based on different ways of calculating the captured variance. In all of them, the captured variance is  added to the variance in  the residuals and normalized by the total sum-of-squares of the data. Therefore, all statistics are expected to be equal to 1}. The considered statistics are defined as:

\begin{itemize}
	\item TotQR: The fraction of total variance computed as: {(QSS+$\trc(\mathbf{E}^\tT\mathbf{E})$)/$\trc(\mathbf{X}^T\mathbf{X})$, where QSS is the sum-of-squares of the diagonal of $\mathbf{B}$ from the QR decomposition of $\mathbf{\hat{T}}=\mathbf{A}\mathbf{B}$.
	\item TotT: The fraction of total variance computed as:
		($\trc(\mathbf{\hat{T}}^\tT\mathbf{\hat{T}})$+$\trc(\mathbf{E}^\tT\mathbf{E})$)/$\trc(\mathbf{X}^\tT\mathbf{X})$, where the first addend in the numerator is the sum-of-squares of the scores.
	\item TotPT: The fraction of total variance computed as:
		($\trc(\mathbf{\hat{P}}\mathbf{\hat{T}}^\tT\mathbf{\hat{T}}\mathbf{\hat{P}}^\tT)$+$\trc(\mathbf{E}^\tT\mathbf{E})$)/$\trc(\mathbf{X}^\tT\mathbf{X})$, where  the first addend in the numerator is the sum-of-squares of the reconstructed data.}
	
\end{itemize}

Numerical results for the non-orthogonal spectra are presented in Table \ref{table:statsa_rest}. The corrected computations lead to the expected value TotPT = 1. However, this is not the case for non-corrected scores {and for other approaches to compute the variance}. As discussed previously, the QR decomposition is, in general, an incorrect way to compute variance, since TotQR is different from 1 in most occasions. {This also holds for TotT}. The amount of variance captured by the orthogonalized methods is the same independently of the statistic used to measured it. 
\begin{table}
	\centering
	\footnotesize
	\begin{tabular}{|c|c|c|c|c|c|c|c|c|}
		\hline
		& \textbf{sPCA} & \textbf{sPCA-Sq} & \textbf{PMD-PD} & \textbf{PMD-O} & \textbf{PMD-M} & \textbf{GPCA-PD} & \textbf{GPCA-M} & \textbf{GPCA-O} \\
		\hline
\hline                                                                                                                                           
\textbf{TotQR} & 1.2730 & 0.9616 & 0.9106 & 1.0000 & 0.9223 & 0.9345 & 0.9345 & 1.0000 \\                                                        
\hline                                                                                                                                           
\textbf{TotT} & 1.7747 & 1.0722 & 1.0000 & 1.0000 & 1.0008 & 1.0000 & 1.0000 & 1.0000 \\                                                         
\hline                                                                                                                                           
\textbf{TotPT} & 2.5494 & 1.1444 & 1.0689 & 1.0000 & 1.0457 & 1.0435 & 1.0435 & 1.0000 \\  
		
		\hline
		\hline
		\hline
\textbf{TotQR*} & 0.8241 & 0.9097 & 0.8541 & 1.0000 & 0.8857 & 0.8947 & 0.8947 & 1.0000 \\                                                        
\hline                                                                                                                                           
\textbf{TotT*} & 0.8606 & 0.9603 & 0.9362 & 1.0000 & 0.9579 & 0.9584 & 0.9584 & 1.0000 \\                                                         
\hline                                                                                                                                           
\textbf{TotPT*} & 1.0000 & 1.0000 & 1.0000 & 1.0000 & 1.0000 & 1.0000 & 1.0000 & 1.0000 \\  
		\hline
	\end{tabular}
\caption{Statistics for the calculation of the fraction of variance captured by the different sparse algorithms for the non-orthogonal spectra example. * Indicates that Eq. (\ref{scores}) is used to compute the scores. In the other cases the scores are computed individually per each component.}
\label{table:statsa_rest}
\end{table}

Figure \ref{fig:Figure_ej100c} shows the effect of the correction in the scores in the Monte Carlo simulation. Without the correction, all methods, except for orthogonalized versions, show variance artifacts in TotPT. This especially affects the simultaneous SPCA. After correction, TotPT is systematically equal to 1, while TotQR and TotT are not, indicating that these statistics are not appropriate to describe the variance captured by the sparse model.

\begin{figure}
	\centering
	\subfigure[scores per-component]{\includegraphics[width=0.3\textwidth]{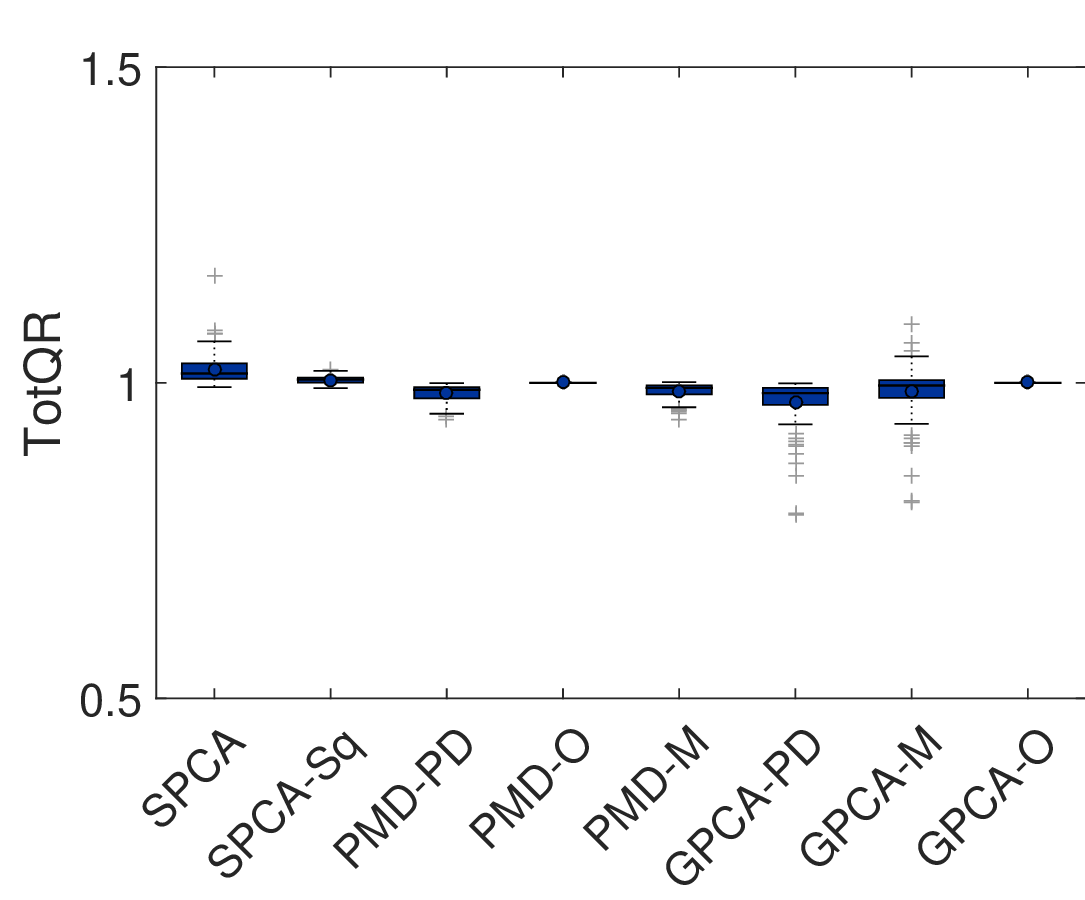} \includegraphics[width=0.3\textwidth]{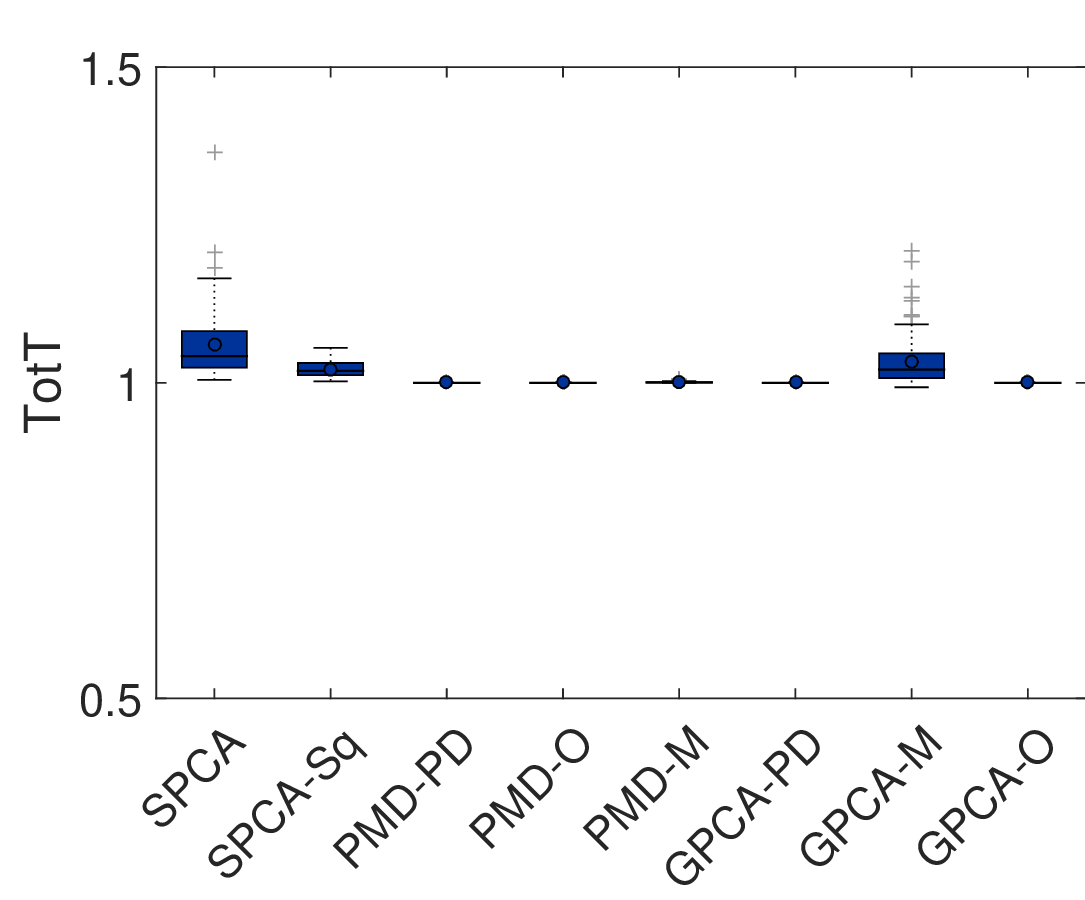}
		\includegraphics[width=0.3\textwidth]{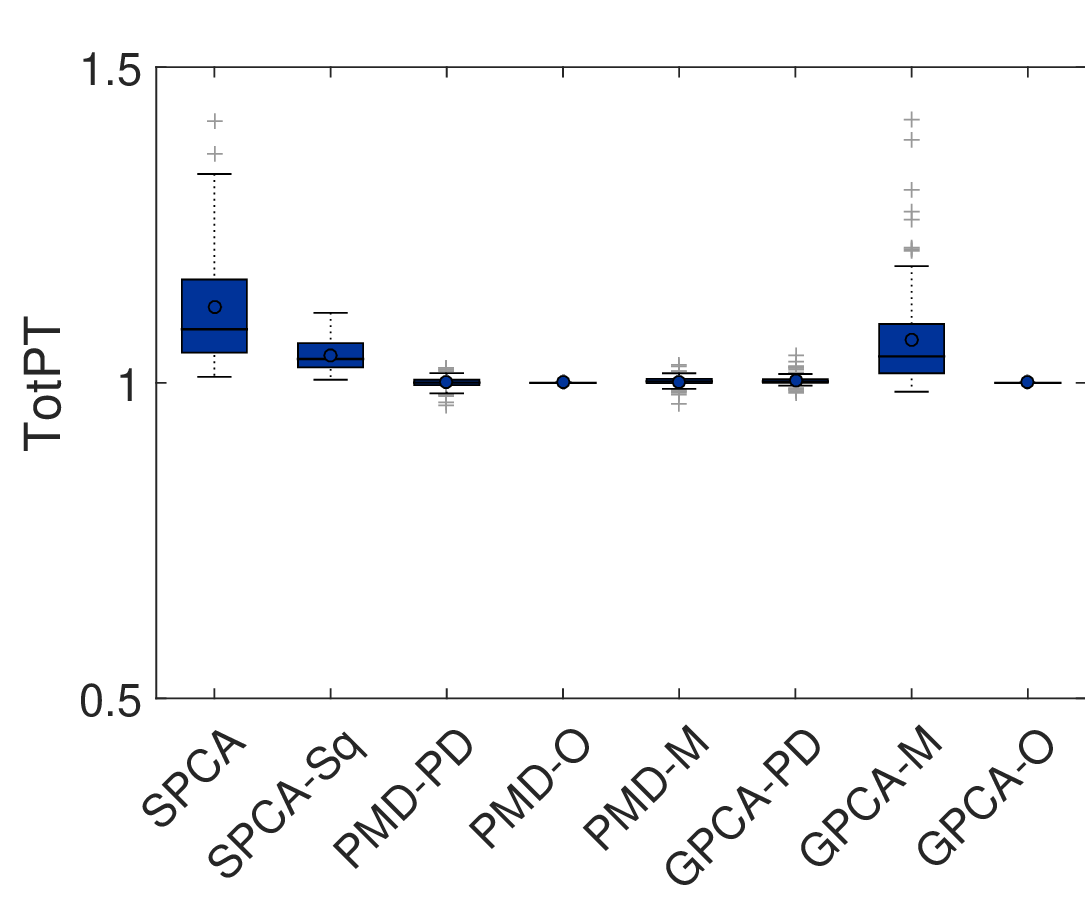}}
	\subfigure[corrected scores]{\includegraphics[width=0.3\textwidth]{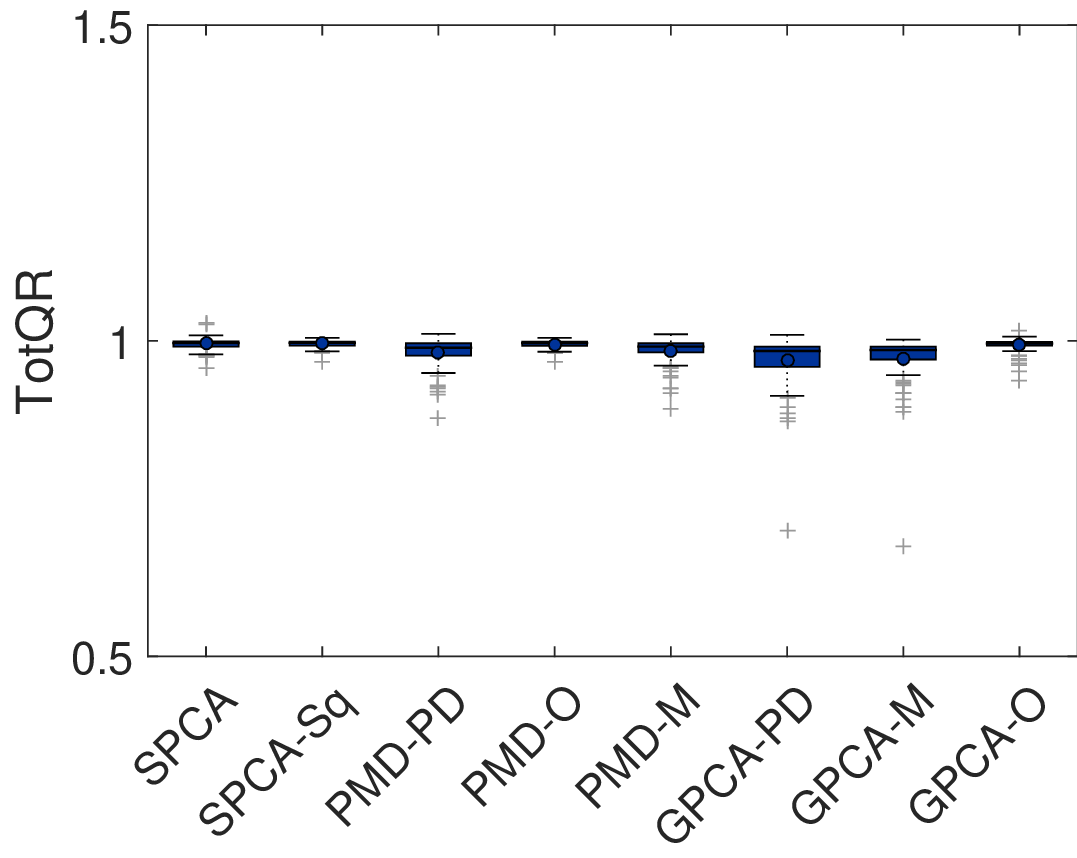}
		\includegraphics[width=0.3\textwidth]{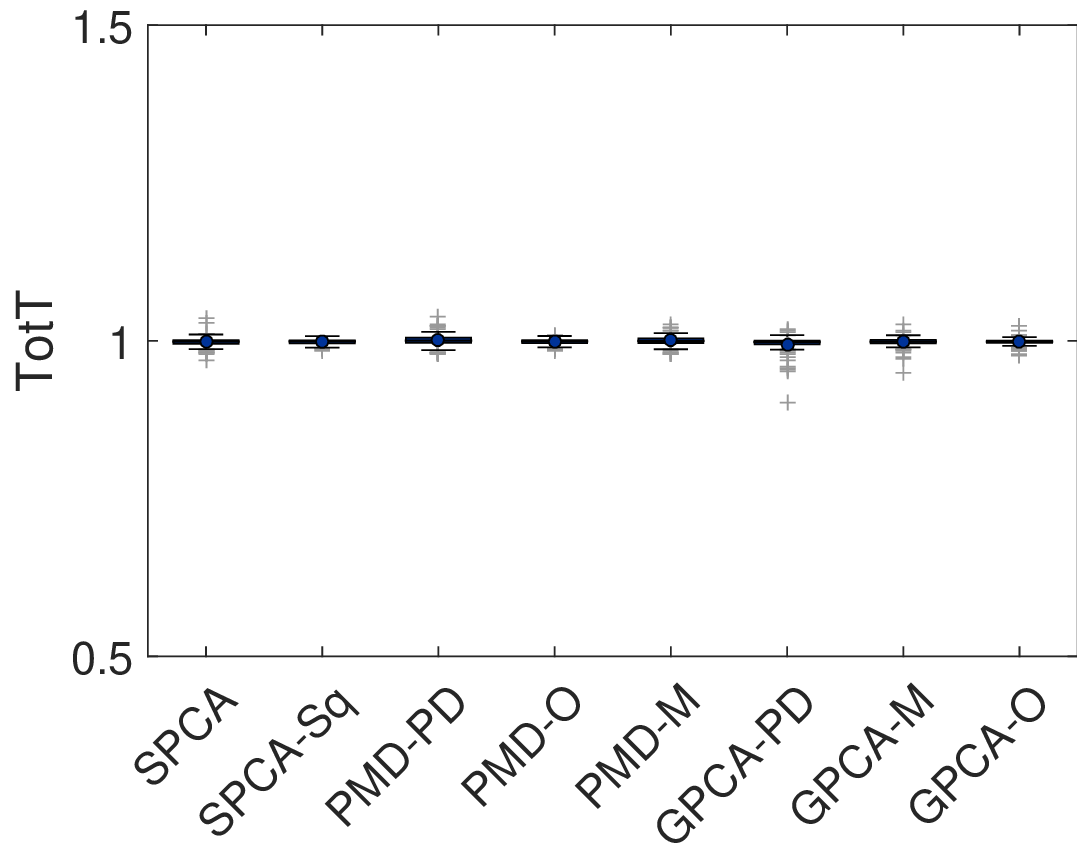}
		\includegraphics[width=0.3\textwidth]{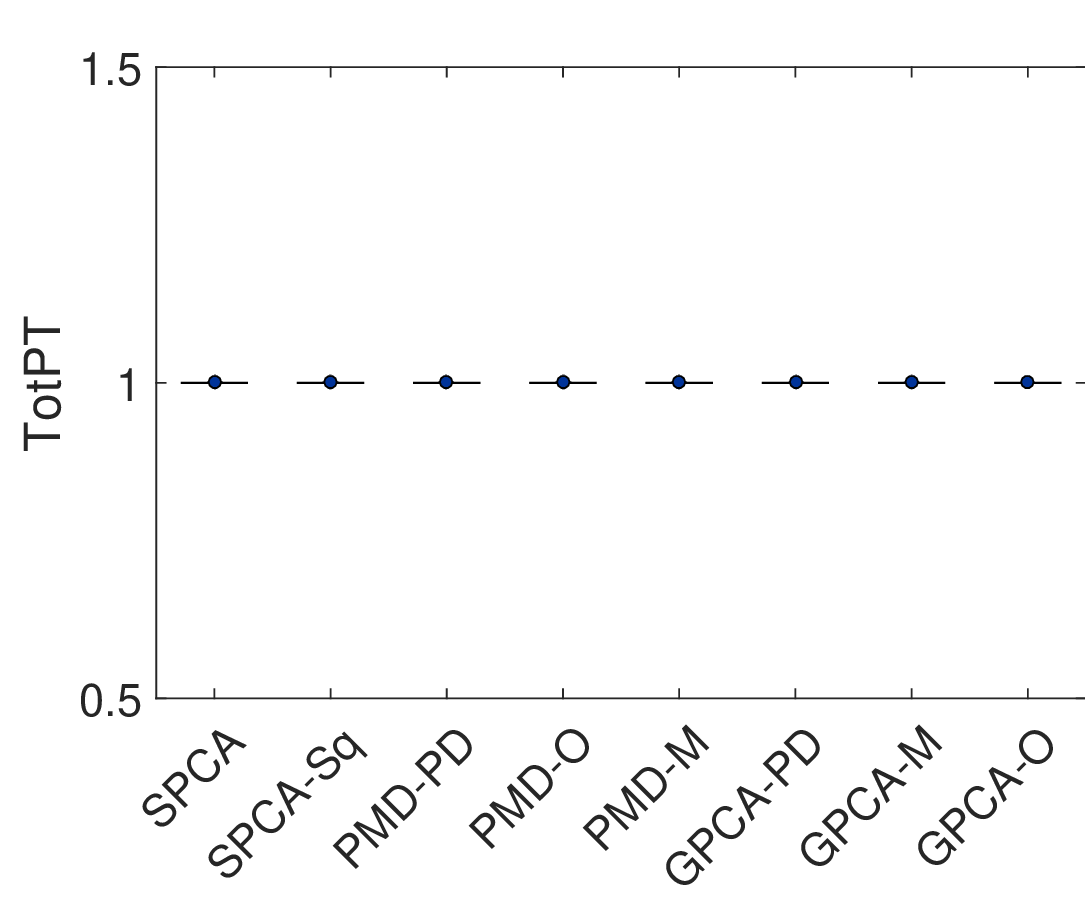}} 
	\caption{Statistics for the calculation of the fraction of variance captured by the different sparse algorithms for the Monte Carlo simulation.}
	\label{fig:Figure_ej100c}
\end{figure}

\section{Conclusion}

{In this paper we reviewed a number of Sparse Principal Component Analysis (sPCA) algorithms and compared them through simulation. 
{We showed that sPCA approaches present disparate and limited performance when modeling sparse data, even in the optimistic case of noise-free data. In a follow-up paper, we discuss the theoretical properties that lead to this problem. The comparative in this and the follow up paper can be useful to make a proper choice of the sPCA algorithm, depending on the practical goal of the analysis. }
	
In this paper we focused on the inherent properties os SPCA, that make it substantially different to PCA and, for this difference, PCA and sparse PCA models must be handled in different ways. In the light of this, we provided corrected equations for quantities such as captured variances, scores and residuals. These corrections are relevant, taking into account that
{	
scores and residuals are central in chemometrics and related disciplines for data visualization and interpretation. Furthermore, explained variance is useful for comparison among model variants, and also to compare sPCA with other modeling approaches.} 

		

}

\appendix

\section{Calculation of the explained variance}
\label{APP:Variance}

For a standard PCA model it holds
\beq
\label{GenericPCAmodel1}
\bX=\bT \bP^\tT + \bE=\hat{\bX}+\bE
\eeq
from which it follows that
\beq
\label{GenericPCAmodel2}
\|\bX\|_F^2=\|\bT \bP^\tT + \bE\|_F^2.
\eeq
Using the general equation $\|\bA\|_F^2=\trc(\bA^\tT\bA)$ it follows that
\beq
\label{GenericPCAmodel3}
\|\bX\|^2=\trc(\bX^\tT\bX)=\trc(\bP\bT^\tT\bT\bP^\tT)+\trc(\bP\bT^\tT\bE)+\trc(\bE^\tT\bT\bP^\tT)+\trc(\bE^\tT\bE)
\eeq
by simply multiplying out the second part of Equation (\ref{GenericPCAmodel2}). If deflation is performed in the score space then $\bT$ are the fixed regressors and a property of the deflation (\textit{i.e.} regression) is that $\bT^\tT\bE=\bE^\tT\bT=0$ and thus
\beq
\label{GenericPCAmodel4}
\|\bX\|^2=\trc(\bX^\tT\bX)=\trc(\bP\bT^\tT\bT\bP^\tT)+\trc(\bE^\tT\bE)=\|\hat{\bX}\|^2+\|\bE\|^2
\eeq
which is a simple split-up of explained variation. If the deflation is performed in the loading space then $\bE\bP=\bP^\tT\bE^\tT=0$ and Equation (\ref{GenericPCAmodel4}) holds since $\trc(\bP\bT^T\bE)=\trc(\bE\bP\bT^\tT)$.\\ 

\noindent Remarks regarding calculating the explained variances:
\begin{enumerate}
	\item The split-up of variation depends on whether Equation (\ref{GenericPCAmodel1}) holds and, additionally, either $\bE\bP=0$ or $\bT^\tT\bE=0$. 
	\item There is no requirement regarding the orthogonality of $\bT$ and/or $\bP$.
	\item There is also
	no requirement of the columns of $\bT$ being in the column space of $\bX$ nor the columns of $\bP$ in the row space of $\bX$.
\end{enumerate}

These equations become simpler if there is orthogonality. If $\bT^\tT\bT=\bI$ then
\beq
\label{GenericPCAmodel5}
\|\bX\|^2=\trc(\bX^\tT\bX)=\trc(\bP\bT^\tT\bT\bP^\tT)+\trc(\bE^\tT\bE)=\trc(\bP\bP^\tT)+\trc(\bE^\tT\bE)=\trc(\bP^\tT\bP)+\trc(\bE^\tT\bE).
\eeq
Similarly, if $\bP^T\bP=\bI$ then 
\beq
\label{GenericPCAmodel6}
\|\bX\|^2=\trc(\bX^\tT\bX)=\trc(\bP\bT^\tT\bT\bP^\tT)+\trc(\bE^\tT\bE)=\trc(\bP^\tT\bP\bT^\tT\bT)+\trc(\bE^\tT\bE)=\trc(\bT^\tT\bT)+\trc(\bE^\tT\bE)
\eeq
which explains why sometimes calculating explained variances using only scores or loadings may work.

 \section*{Acknowledgement}
\label{sec:Acknowledgments}
This work is partly  supported by the Spanish Ministry of Economy and Competitiveness and FEDER funds through  project TIN2017-83494-R and the "Plan Propio de la Universidad de Granada".

\bibliographystyle{ama}
\bibliography{Bibliography}

\end{document}